\documentclass[10pt,journal,compsoc]{IEEEtran}

\usepackage{amsmath}
\usepackage{diagbox}
\usepackage{graphics}
\usepackage{epsfig}
\usepackage{threeparttable}
\usepackage{xspace}
\usepackage{siunitx}
\usepackage{graphicx}
\usepackage{amssymb}
\usepackage{threeparttable}
\usepackage{color}
\usepackage[normalem]{ulem}
\usepackage{multirow}
\usepackage{float}
\usepackage{amsfonts}
\usepackage{bm}
\usepackage{bbm}
\usepackage{array}
\usepackage[table]{xcolor}
\usepackage{colortbl}
\usepackage{diagbox}
\usepackage{rotating}
\usepackage{booktabs}
\usepackage{overpic}
\usepackage{enumitem}
\usepackage{colortbl}
\usepackage{algorithm}
\usepackage{pythonhighlight}
\usepackage{algorithmic}
\usepackage{cite}
\usepackage{arydshln}
\usepackage[caption=false, font=scriptsize]{subfig}
\usepackage[pagebackref=false,breaklinks=true,colorlinks,bookmarks=false]{hyperref}
\usepackage{cleveref}

\definecolor{mygray}{gray}{.85}
\definecolor{mygray1}{gray}{.7}
\definecolor{mygray2}{gray}{.93}

\makeatletter
\newcommand{\thickhline}{%
	\noalign {\ifnum 0=`}\fi \hrule height 1pt
	\futurelet \reserved@a \@xhline
}
\makeatother

\makeatletter
\DeclareRobustCommand\onedot{\futurelet\@let@token\@onedot}
\def\@onedot{\ifx\@let@token.\else.\null\fi\xspace}

\def\eg{\emph{e.g}\onedot} 
\def\ie{\emph{i.e}\onedot} 
 
\def\etc{\emph{etc}\onedot}

\makeatother

\usepackage{pifont}
\usepackage{bbding}
\usepackage{fontawesome}

\newcommand{\cmark}{\ding{51}}
\newcommand{\xmark}{\ding{55}}

\definecolor{mygray1}{gray}{.4}
\definecolor{mywarning}{RGB}{233,144,61}
\definecolor{myred}{RGB}{192,0,0}
\definecolor{myyellow}{RGB}{255,192,0}
\definecolor{myorange}{RGB}{243,225,214}
\definecolor{bblue}{RGB}{0,150,230}
\definecolor{mygray}{gray}{.9}
\definecolor{myy}{RGB}{126,95,0}
\definecolor{ggray}{RGB}{127,127,127}
\definecolor{mygreen}{RGB}{93,174,86}
\definecolor{myblue}{RGB}{0,114,188}
\definecolor{darkgreen}{rgb}{0.0, 0.5, 0.0}
\definecolor{demphcolor}{RGB}{100,100,100}
\definecolor{rowblue}{RGB}{198,234,251}   
\definecolor{rowgreen}{RGB}{209,239,191}  
\definecolor{rowgray}{RGB}{245,245,245}   
\definecolor{roworange}{RGB}{243,225,214} 

\begin{document}
	\title{Dynamic in Static:\\ Hybrid Visual Correspondence for\\ Self-Supervised Video Object Segmentation}
	\author{Gensheng Pei, Yazhou Yao, Jianbo Jiao, Wenguan Wang, Liqiang Nie, and Jinhui Tang
		\IEEEcompsocitemizethanks{
			\IEEEcompsocthanksitem G. Pei, Y. Yao, and J. Tang are with Nanjing University of Science and Technology. (Email: \{peigsh, yazhou.yao, jinhuitang\}@njust.eud.cn)
			\IEEEcompsocthanksitem W. Wang is with Zhejiang University. (Email: wenguanwang.ai@gmail.com)
			\IEEEcompsocthanksitem J. Jiao is with University of Birmingham. (Email: j.jiao@bham.ac.uk)
			\IEEEcompsocthanksitem L. Nie is with Harbin Institute of Technology (Shenzhen). (Email: nieliqiang@gmail.com)			
		}
	}


\IEEEtitleabstractindextext{
\begin{abstract}
Conventional video object segmentation (VOS) methods usually necessitate a substantial volume of pixel-level annotated video data for fully supervised learning. In this paper, we present HVC, a \textbf{h}ybrid static-dynamic \textbf{v}isual \textbf{c}orrespondence framework for self-supervised VOS. HVC extracts pseudo-dynamic signals from static images, enabling an efficient and scalable VOS model. Our approach utilizes a minimalist fully-convolutional architecture to capture static-dynamic visual correspondence in image-cropped views. To achieve this objective, we present a unified self-supervised approach to learn visual representations of static-dynamic feature similarity. Firstly, we establish static correspondence by utilizing a priori coordinate information between cropped views to guide the formation of consistent static feature representations. Subsequently, we devise a concise convolutional layer to capture the forward / backward pseudo-dynamic signals between two views, serving as cues for dynamic representations. Finally, we propose a hybrid visual correspondence loss to learn joint static and dynamic consistency representations. Our approach, without bells and whistles, necessitates only one training session using static image data, significantly reducing memory consumption ($\sim$16GB) and training time ($\sim$\textbf{2h}). Moreover, HVC achieves state-of-the-art performance in several self-supervised VOS benchmarks and additional video label propagation tasks.
The source code and models are available at: \href{https://github.com/NUST-Machine-Intelligence-Laboratory/HVC}{https://github.com/NUST-Machine-Intelligence-Laboratory/HVC}.
\end{abstract}
}

\maketitle

\IEEEdisplaynontitleabstractindextext
\IEEEpeerreviewmaketitle

\IEEEraisesectionheading{
\section{Introduction}\label{sec:introduction}}
\IEEEPARstart{V}{ideo} object segmentation (VOS) aims to separate and track objects of interest in videos, a task traditionally dependent on labor-intensive annotated data. The move towards self-supervised methods enables learning from unlabeled data, cutting the need for pixel-level annotations. Among these methods, visual correspondence learning has emerged as a particularly promising direction.
Accordingly, self-supervised visual correspondence learning is introduced, which learns correspondence from raw videos without any external human supervision~\cite{jabri2020space,lai2020mast,xu2021rethinking,li2022liir,son2022contrastive,dense2021,li2023unified,wang2021contrastive,lai2019self,li2023spatial,yao2021non,zhao2021modelling,li2022pixels,PerSAM}.
With the cooperation of correspondence learning across space and time, several downstream tasks, such as optical flow estimation~\cite{yin2018geonet,ye2021motion,xu2022gmflow}, video object tracking~\cite{yan2022towards,cheng2023segment,cao2023observation}, and VOS~\cite{vondrick2018tracking,zhou2023survey,miles2023mobilevos,perazzi2016benchmark}, have made considerable progress.

\begin{figure}[t]
    \begin{center}
        \includegraphics[width=1.0\linewidth]{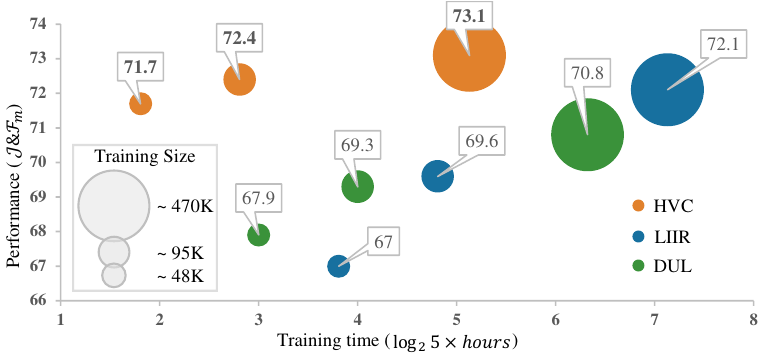}
        \put(-26.3,49.95){\scriptsize (\textbf{ours})}
        \put(-26.3,39.75){\scriptsize\cite{li2022liir}}
        \put(-26.3,29.90){\scriptsize\cite{dense2021}}
    \end{center}
    \vspace{-8pt}
    \caption{\textbf{Self-supervised VOS performance comparison} on DAVIS$_{17}$ \cite{pont2017}~\texttt{val-set}. We adopt the same hardware device and training data~\cite{xu2018youtube} for a fair comparison. Bubble size indicates the size of training images.Taking less training time, our model outperforms state-of-the-art methods across all scales of data volume. HVC trained with 95K images achieves comparable performance to LIIR~\cite{li2022liir} delivered with 470K ones.}\label{fig:1}
    \vspace{-8pt}
\end{figure}

To design self-supervision signals, recent studies have employed various pretext tasks, including colorization~\cite{lai2019self,lai2020mast}, cycle-consistency constraints~\cite{wang2019learningcorr,li2019joint,lu2020learning}, instance discrimination~\cite{wu2018unsupervised,jeon2021mining}, or image-level similarity~\cite{xu2021rethinking}. Promising developments have been achieved by these approaches.
Yet, a noise-based model~\cite{vondrick2018tracking} for pretext tasks could lead to local optima and overfitting~\cite{son2022contrastive}. To minimize noisy label effects, several studies~\cite{jeon2021mining,xu2021rethinking} have proposed dynamic correspondence learning, which is based on contrastive matching and draws inspiration from contrastive learning~\cite{chen2020simple}.
This approach involves contrasting the affinity of matched pixel pairs (\textit{positive samples}) with that of unrelated pixel pairs (\textit{negative samples}). 
Despite these efforts, there remains significant uncertainty in distinguishing between positive and negative samples.
In response, recent works~\cite{jabri2020space,dense2021,son2022contrastive} have presented more accurate positive and negative sample mining techniques to address this challenge.
Usually, spatio-temporal neighboring pixels are defined as positive pairs but still suffer from noise when the inter-frame feature differences are severe.
To alleviate the aforementioned challenge, a set of solutions \cite{grill2020bootstrap,chen2020simsiam,vangansbeke2021revisiting,xie2021propagate} directly utilize positive sample pairs for self-supervised feature representation learning, eliminating the need to grapple with the complexities of distinguishing between positive and negative samples.
For various downstream tasks (\eg, classification~\cite{ConvNeXtV2,fang2023eva}, detection~\cite{caron2020unsupervised,xu2021end}, and segmentation~\cite{xie2021segformer,wang2021pyramid}), the image-level global feature representations learned from this scheme still require dataset-specific fine-tuning and cannot be directly applied to dense video segmentation.
Nevertheless, they undeniably pave the way towards self-supervised feature representation without using negative samples.

\begin{figure*}[t]
    \begin{center}
        \includegraphics[width=1.0\linewidth]{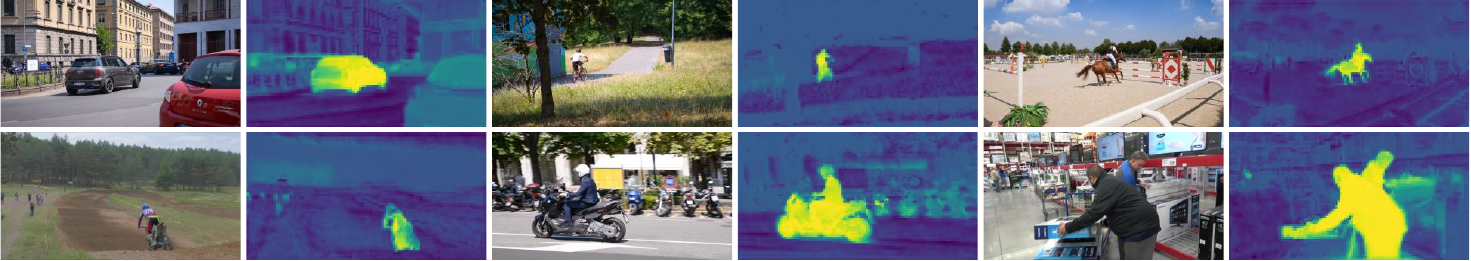}
    \end{center}
    \vspace{-6pt}
    \caption{\textbf{Learned representation visualization from HVC without any supervision.} Our proposed self-supervised hybrid visual correspondence learning highlights salient regions, suggesting the suitability of the HVC-learned representations for dense tasks such as video object segmentation.}\label{fig:2}
    \vspace{-6pt}
\end{figure*}

Existing self-supervised VOS approaches~\cite{jabri2020space,lai2020mast,xu2021rethinking,li2022liir,son2022contrastive,dense2021,li2023unified} focus on learning spatio-temporal feature representations from videos for dense segmentation tasks. However, one notable drawback of video-based correspondence learning is the relatively higher training costs compared to static images. Recently, some studies~\cite{vangansbeke2021revisiting,caron2021emerging,CrOC} have revisited MoCo~\cite{he2020momentum} and proposed a multi-scale cropping strategy. This strategy enables models to learn static structure representations from images, making them applicable for VOS without the need for weight fine-tuning.
However, it is worth noting that these methods~\cite{vangansbeke2021revisiting,caron2021emerging,CrOC} are built on the Vision Transformer (ViT) backbone approach~\cite{dosovitskiy2020image}, resulting in persistently high computational costs for training, and they do not effectively address the challenge of dynamic consistency.

In light of this context, one important question naturally arises: \textit{Is it possible to learn spatio-temporal visual correspondence from static images only, thereby achieving self-supervised VOS in a more efficient manner?}

To address this question, we propose a method termed \textit{hybrid visual correspondence} (HVC), which integrates \textit{static} and \textit{dynamic} visual correspondence learning from images into an elegant and efficient framework for self-supervised VOS.
Departing from the conventional video data-driven methods, HVC eliminates the complexities of cross-frame reconstruction while attaining superior performance and resource efficiency.
By exclusively leveraging static images, our approach exploits a vast reservoir of unlabeled images, showcasing diverse scenes, objects, and perspectives. These images act as an extensive repository of visual data, propelling the development of robust and widely applicable VOS models.
However, a key challenge lies in the fact that unlike video data, which inherently encompasses dynamic information, static images lack the essential dynamic signals required for self-supervised VOS pretext tasks.
It is not trivial to inject dynamic signals into static images.

Motivated by the aforementioned discussions, we propose a pseudo-dynamic learning strategy aimed at extracting pseudo-dynamic information from static images while incorporating their inherent static consistency present in these images.
To achieve this, building upon a fully convolutional architecture, our model learns \textbf{i)} static consistency across image-cropped views and \textbf{ii)} dynamic consistency across pseudo-dynamic signals.
Static consistency is attained by modeling similarities between image-cropped views, thereby bringing pixel pairs closer together in overlapping regions.
We introduce pseudo-dynamic signals from static images to ensure dynamic consistency.
This is accomplished through a meticulously designed pseudo-dynamic generation network, which determines the offset vector for each pixel in the cropped views of images with overlapping regions.
To create positive masks, we leverage the determined coordinate relationships between cropped views of the images. These masks play a vital role in identifying the regions containing deterministic positive pixel pairs in both the image-cropped views (\textit{static}) and the pseudo-dynamic signals (\textit{dynamic}). This process enables the construction of hybrid visual correspondence more effectively.
Under a self-supervised paradigm, our approach ensures that the learned representations from only images achieve static-dynamic consistency, thereby fulfilling the stated aim.

We conduct extensive experiments and comprehensive ablation studies to validate the proposed approach. For self-supervised VOS, HVC demonstrates superior performance compared to state-of-the-arts, achieving $\mathcal{J}\&\mathcal{F}_{m}$ gains of \textbf{1.3\%}, \textbf{1.0\%}, \textbf{1.8\%}, \textbf{2.0\%}, and \textbf{6.7\%} on the DAVIS$_{16}$~\cite{perazzi2016benchmark} \texttt{val-set}, DAVIS$_{17}$~\cite{pont2017} \texttt{val-set}, DAVIS$_{17}$ \texttt{test-set}, YouTube-VOS$_{18}$~\cite{xu2018youtube} \texttt{val-set}, and YouTube-VOS$_{19}$~\cite{yang2019video} \texttt{val-set}, respectively. HVC sets new state-of-the-art (\textbf{26.3\%} $\mathcal{J}_{m}$) for the challenging VOST~\cite{tokmakov2023breaking} benchmark. Furthermore, our approach yields competitive results in two additional tasks involving video label propagation, including part segmentation (\textbf{44.6\%} mIoU on VIP~\cite{zhou2018adaptive}) and pose tracking (\textbf{61.7\%} PCK@0.1 and \textbf{82.8\%} PCK@0.2 on JHMDB~\cite{jhuang2013towards}).

To sum up, the main contributions of this work include:
\begin{itemize}
    \item[1)] We propose a new self-supervised approach termed as hybrid visual correspondence (HVC), which effectively combines static and dynamic cues to tackle the challenge of label-free VOS. As shown in Fig.~\ref{fig:1}, HVC outperforms state-of-the-arts, achieving superior results while requiring notably less training data and time.
    \item[2)] We introduce static and dynamic correspondence schemes that ensure the learned representations uphold static-dynamic consistency, a critical factor for VOS. Fig.~\ref{fig:2} shows examples of mask propagation using HVC.
    \item[3)] Our approach provides an elegant and efficient solution to liberate self-supervised VOS from its reliance on video data. By using only \textbf{static images} as training data, HVC surpasses the existing self-supervised strategies that depend on video data. Remarkably, HVC achieves exceptional performance utilizing a mere \textbf{16GB} GPU memory and a brief training duration of only \textbf{2 hours}.
\end{itemize}

\section{Related Work}
\subsection{Fully-Supervised Video Object Segmentation}
VOS aims to segment universal foreground objects in video sequences, regardless of their semantic classes. In the inference stage, VOS is broadly classified into one-shot (or semi-supervised) VOS~\cite{oh2019video,caelles2017one,cheng2022xmem,zhang2023boosting} and zero-shot (or unsupervised) VOS~\cite{pei2022hierarchical,wang2019learning,zhang2021deep,HCPN,HGPU}, depending on whether the object masks of the first frame target are provided. In the fully supervised VOS setting, inter-frame correspondence associations are established by learning feature representations to propagate semantic labels. Most existing approaches, whether based on matching~\cite{park2022per,oh2019video,seong2021hierarchical,seong2022video}, propagation~\cite{cheng2021modular,perazzi2017learning,xu2022reliable,yang2022decoupling}, or online fine-tuning~\cite{caelles2017one,voigtlaender2017online,maninis2018video}, require large amounts of pixel-level annotated video datasets for supervised model training. These approaches typically use ImageNet~\cite{deng2009imagenet} pre-trained backbones and fine-tune models on target datasets (\eg, YouTube-VOS~\cite{xu2018youtube} and DAVIS~\cite{perazzi2016benchmark}) to achieve satisfactory performance.
Unfortunately, this heavy reliance on expensive pixel-level annotated videos poses a significant limitation for fully supervised VOS methods.

Unlike the aforementioned approaches based on annotated videos, MaskTrack~\cite{perazzi2017learning} stands out as one of the pioneering attempts to address VOS using only static images. It aims to achieve high-precision video segmentation by leveraging static image data. However, it is worth noting that MaskTrack~\cite{perazzi2017learning} is a fully supervised approach, which still requires many ground truth annotations for training.
The quest for self-supervised methods that can learn VOS without relying on fully annotated videos has emerged as an active area of research.
In this paper, we present a self-supervised VOS method that integrates both static and dynamic visual correspondence. Our approach addresses the challenge posed by limited labeled data in the VOS setting. By incorporating both static and dynamic information, HVC aims to enhance the accuracy and robustness of the segmentation process. This joint scheme overcomes the limitations posed by insufficient labeled data, offering a promising solution for self-supervised VOS.

\subsection{Self-Supervised Video Object Segmentation}
Self-supervised learning uses unlabeled data by constructing pretext tasks to learn visual representations.
Typically, self-supervised visual representation learning does not focus on the final performance of the pretext tasks.
Many works concentrate on discriminative methods~\cite{tian2020contrastive,chen2020simple,wu2018unsupervised,he2020momentum,oord2018representation,chen2020improved} towards instance-level classification. In this approach, each image is treated as a separate category, and the core idea is to bring feature embeddings of the same image view closer while pushing features of different images further apart. 
Recent works~\cite{grill2020bootstrap,chen2020simsiam,caron2021emerging,vangansbeke2021revisiting} have demonstrated that self-supervised learning can acquire feature representations without instance discrimination.
Notably, \cite{caron2021emerging} attempts to address self-supervised VOS tasks by leveraging the powerful representation capabilities of ViT~\cite{dosovitskiy2020image}.
IKE~\cite{haller2021iterative} introduces a novel self-supervised learning framework that uses iterative knowledge exchange between deep learning and space-time spectral clustering, enhanced by longer motion chains, to achieve unsupervised object segmentation in videos.
In recent years, the emergence of large-scale visual-language models like CLIP~\cite{CLIP} has revealed their significant potential, leading to the development of language-guided self-supervised VOS methods~\cite{MaskCLIP,CLIP-S4}.

Recently, some studies~\cite{xie2021propagate,CrOC,hu2022sfc} have revealed that constructing self-supervised pretext tasks using unlabeled image-level data~\cite{lin2014microsoft,everingham2010pascal} can be extended to dense VOS scenarios as well.
For instance, PixPro~\cite{xie2021propagate} leverages the consistency of pixel-level information across different augmented views of the same image. By training the model to predict pixel-level relationships and enforcing consistency constraints, the framework learns powerful visual representations. Building upon this idea, subsequent studies, SFC~\cite{hu2022sfc} and CrOC~\cite{CrOC}, have made substantial advancements in VOS by implementing a self-supervised training strategy for static image cross-views, closely resembling the method introduced by PixPro~\cite{xie2021propagate}. These approaches have demonstrated noteworthy performance improvements in the VOS task. Although the cross-view visual representation can achieve static consistency, the lack of dynamic consistency is a common problem. We propose a hybrid visual correspondence approach that integrates both static and dynamic signals into the self-supervised learning process. Specifically, we inject dynamic signals into images to embed static and dynamic consistency within the original image-level visual correspondence learning. As a result, our method outperforms existing approaches trained on video datasets and achieves state-of-the-art performance across multiple VOS benchmark datasets.

\subsection{Space-Time Correspondence Learning}
Extracting inter-frame correlation has been a prominent and essential objective in computer vision, as it plays a vital role in various video applications, including optical flow estimation and object tracking. Recent methods~\cite{jabri2020space,lai2020mast,xu2021rethinking,li2022liir,son2022contrastive,dense2021,li2023unified,wang2021contrastive,lai2019self,li2023spatial,zhao2021modelling,li2022pixels} have made significant advancements in space-time video correspondence learning by leveraging self-supervised approaches.
Color-consistency methods~\cite{lai2019self,lai2020mast,vondrick2018tracking}, drawing inspiration from video colorization, facilitate dynamic correspondence learning by comparing the affinity of current frame colors with the propagated future frame colors.
Another line of approach utilizes circle-consistency constraint as their self-supervisory signals~\cite{li2019joint,jabri2020space,dense2021,son2022contrastive,lu2020learning}.
These methods aim to acquire static-dynamic feature representations by tracking both forward and backward pixels or regions among neighboring frames.
Furthermore, certain methods~\cite{ponimatkin2022simple,ding2022motion,yang2021self,liu2021emergence,haller2019spacetime} leverage optical flow~\cite{ye2021motion,yin2018geonet} to establish dynamic consistency constraints and learn features for dynamic correspondence.
These approaches have significantly advanced self-supervised correspondence learning for video object segmentation.
However, these solutions are exclusively designed for video sequences and heavily rely on inter-frame dynamic cues to guide self-supervised pretext tasks.
In contrast, our hybrid visual correspondence learning approach does not require the acquisition of dynamic signals from video data. 
Instead, we subtly inject dynamic signals into the static images through a simple cropping operation.
Significantly, the proposed approach outperforms the majority of video-based visual correspondence schemes.

\begin{figure*}[t]
    \begin{center}
        \includegraphics[width=1.0\linewidth]{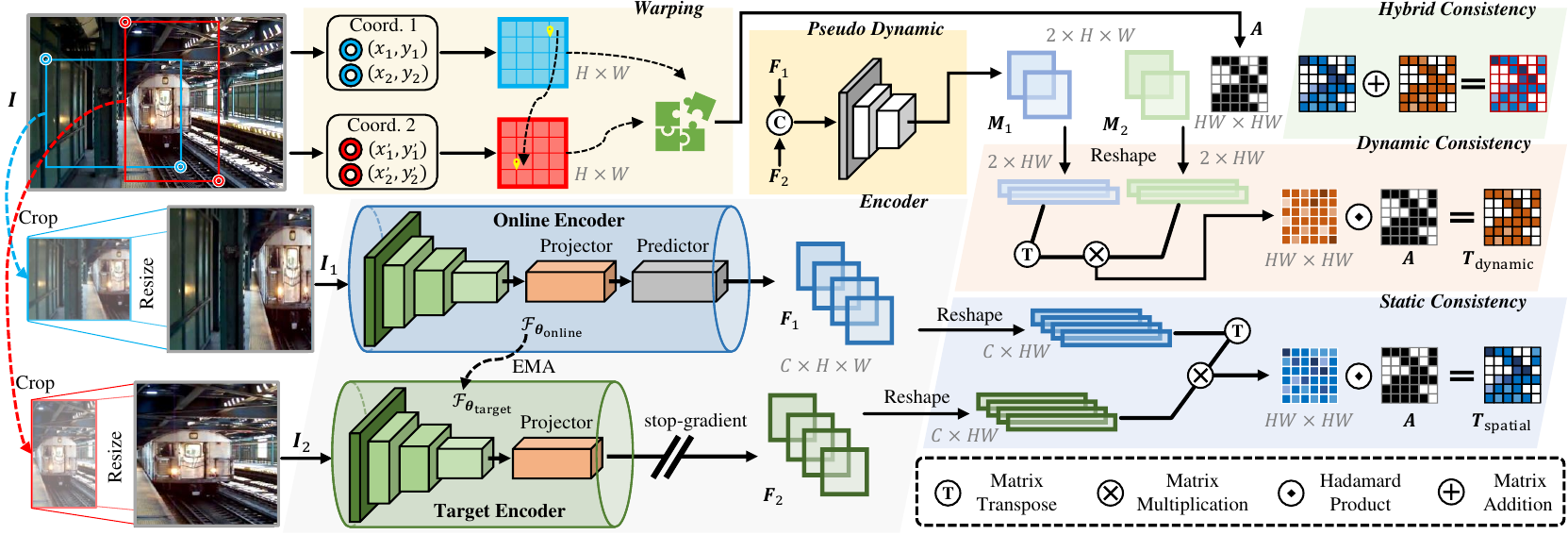}
        \put(-232.00,99.50){\scriptsize Eq.~\eqref{eq:1}}
        \put(-225.00,156.50){\scriptsize Eq.~\eqref{eq:3}}
        \put(-306.00,162.50){\scriptsize Eq.~\eqref{eq:8}}
        \put(-35.00,162.50){\scriptsize Eq.~\eqref{eq:6}}
        \put(-35.00,66.50){\scriptsize Eq.~\eqref{eq:5}}
        \put(-35.00,118.50){\scriptsize Eq.~\eqref{eq:5}}
        \put(-131.50,164.00){\scriptsize Eq.~\eqref{eq:4}}
    \end{center}
    \vspace{-7pt}
    \caption{\textbf{Architecture of the proposed HVC.} Given an image $\bm{I}$, its pair of views $\bm{I}_{1}$ and $\bm{I}_{2}$ are obtained by crop-resize transformation. Two views are fed to the online and target networks, $\mathcal{F}_{\bm{\theta}_{\rm{online}}}$ and $\mathcal{F}_{\bm{\theta}_{\rm{target}}}$ (\S\ref{sec:3.1}). The online network contains the additional projector and predictor heads to acquire feature maps $\bm{F}_{1}$ and the target network appends only projector heads to receive feature maps $\bm{F}_{2}$. The pseudo-dynamic signal generation module receives $\bm{F}_{1}$ and $\bm{F}_{2}$ and outputs forward and backward pseudo-dynamic signals, $\bm{M}_{1}$ and $\bm{M}_{2}$. Inter-feature and inter-dynamic similarities are combined with a positive sample mask to yield hybrid similarity (taking the negative, \ie, final loss, see \S\ref{sec:3.2} for details).}\label{fig:ivr}
\end{figure*}
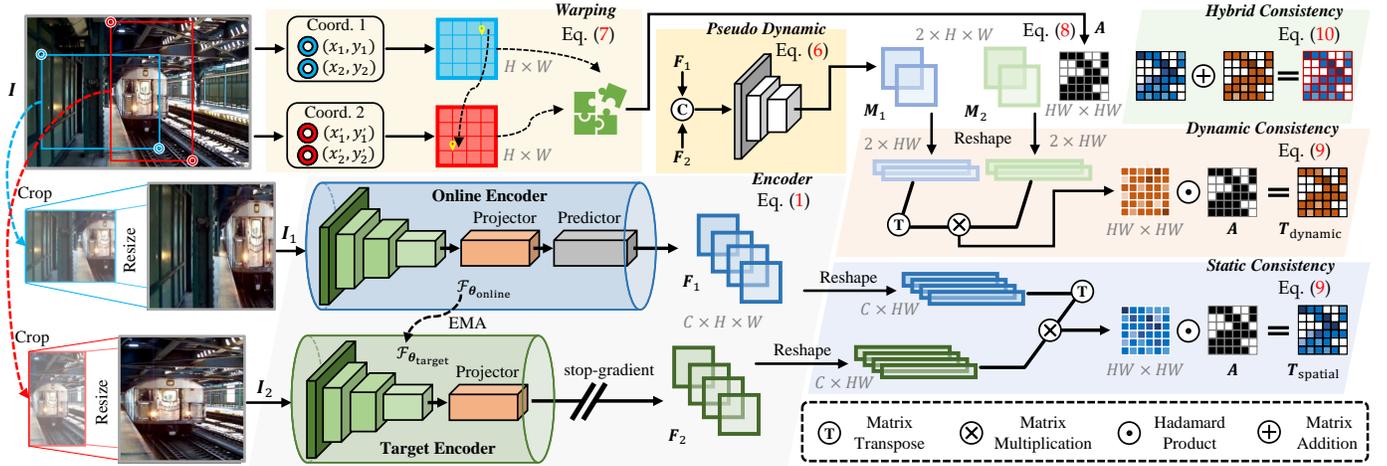

\section{The Proposed Method}
In this study, we introduce a hybrid visual correspondence framework designed to learn both static and dynamic correspondence solely from image data without any human annotations. In the following sections, we will introduce the overview of our framework (\S\ref{sec:3.1}) illustrated in Fig.~\ref{fig:ivr}, then elaborate on the proposed method (\S\ref{sec:3.2}). Additionally, we will delve into the implementation details (\S\ref{sec:3.3}) of HVC.

\subsection{Framework}\label{sec:3.1}
In this study, we present an elegant and efficient visual correspondence learning method for video object segmentation, using label-free static images. Unlike conventional video correspondence learning methods that rely on video sequences, our approach aims to learn feature representations from static images and applies them to a self-supervised VOS task. Consequently, our model is trained only on static images. We show the architecture of the proposed method in Fig.~\ref{fig:ivr} and provide the PyTorch-like style pseudo-code in Algorithm~\ref{alg:HVC}.

Given an image $\bm{I}$, the pair of views $\bm{I}_1, \bm{I}_2\in\mathbb{R}^{{3}\times{h}\times{w}}$ is obtained by randomly cropping transformation~\cite{he2020momentum}.
HVC feeds image-cropped views into the online $\mathcal{F}_{\bm{\theta}_{\rm{online}}}$ and target $\mathcal{F}_{\bm{\theta}_{\rm{target}}}$ encoder networks. Both encoders comprise a backbone and a projector head. In particular, the online encoder brings in an additional predictor head to create an asymmetric structure between the online and target encoder networks, following the generalized methodology~\cite{xie2021self}. The projector and predictor heads are constructed with a space projection module, which consists of three consecutive 1×1 \texttt{Conv} layers (in this work, we set the output channels to 256, see \S\ref{sec:abs}). Here, a \texttt{BN} layer and a \texttt{ReLU} layer are inserted between neighbouring \texttt{Conv} layers to generate feature maps with a specific spatial resolution (set to $32\times{32}$ in this work). 

Following~\cite{caron2021emerging,he2020momentum}, the online encoder is updated by gradient, and we apply the stop-gradient operator to the target encoder. The target encoder parameters $\bm{\theta}_{\rm{target}}$ are updated in each training iteration by an exponential moving average (EMA) of the online encoder momentum parameters $\bm{\theta}_{\rm{online}}$. Hence, the whole encoder architecture is:
\begin{equation}
    \begin{aligned}\label{eq:1}
    \bm{\theta}_{\rm{target}}\leftarrow{m\bm{\theta}_{\rm{target}}+(1-m)\bm{\theta}_{\rm{online}}}, m\in{[0,1]},
    \end{aligned}
\end{equation}
where $m$ is a momentum coefficient, and the setting of its momentum update strategy is detailed in \S\ref{sec:4}. 

To prevent a cumbersome training strategy, we only employ positive pairs to perform feature embeddings without negative pairs, which has been proven effective by recent works~\cite{grill2020bootstrap,chen2020simsiam,caron2021emerging}.
The feature maps of the two cropped views from one image are computed by $\mathcal{F}_{\bm{\theta}_{\rm{online}}}$ and $\mathcal{F}_{\bm{\theta}_{\rm{target}}}$ and normalized by $l_{2}$ normalization as:
\begin{equation}
    \begin{aligned}\label{eq:flow}
    \bm{F}_1,\bm{F}_2=l_{2}(\mathcal{F}_{\bm{\theta}_{\rm{online}}}(\bm{I}_1)),l_{2}(\mathcal{F}_{\bm{\theta}_{\rm{target}}}(\bm{I}_2))\in\mathbb{R}^{{C}\times{H}\times{W}}.
    \end{aligned}
\end{equation}

We apply a simple contrastive loss to learn the feature representations and construct the baseline model for visual correspondence learning, \ie, minimizing the distance (measured by cosine similarity) between the projected features. The loss function is defined as:
\begin{equation}
    \begin{aligned}\label{eq:2}
    \mathcal{L} = -\frac{<\bm{F}_1, \bm{F}_2>}{||\bm{F}_1||_{2}~||\bm{F}_2||_{2}},
    \end{aligned}
\end{equation}
where $<\!\cdot, \cdot\!>$ denotes inner product. The feature maps of $\bm{F}_1$ and $\bm{F}_2$ do not match directly since they come from two different views after cropping and resizing. 
Therefore, the main issue addressed in Eq.~\eqref{eq:2} is finding the optimal positive pairs between $\bm{F}_1$ and $\bm{F}_2$.
In \S\ref{sec:3.2.2}, we will elaborate on this in detail and present a simple solution to achieve image-based correspondence learning for self-supervised VOS.

\begin{algorithm}[t]
   \caption{HVC PyTorch-like pseudocode.}
   \label{alg:HVC}
    \definecolor{codeblue}{rgb}{0.25,0.5,0.5}
    \definecolor{codered}{rgb}{0.80,0.25,0.25}
    \lstset{
      basicstyle=\fontsize{7.6pt}{7.6pt}\ttfamily\bfseries,
      commentstyle=\fontsize{7.6pt}{7.6pt}\color{codeblue},
      keywordstyle=\fontsize{7.6pt}{7.6pt}\color{codered},
    }
\begin{lstlisting}[language=python]
# fo, ft: online and target encoder networks
# r: positive radius
# m: momentum coefficient
# pseudo: pseudo dynamic network
# load two crop views and coordinates
for v1, v2, c1, c2 in loader: 
    # online output BxCxHxW
    o1, o2 = fo(v1), fo(v2)
    # target output BxCxHxW
    t1, t2 = ft(v1), ft(v2)
    # compute hybrid loss
    loss = hybrid_loss(o1, t2, c1, c2, r) + \
           hybrid_loss(o2, t1, c2, c1, r)
    # back-propagate
    loss.backward() 
    # update online and target parameters
    update(fo)
    ft = m * ft + (1.0 - m) * fo
    # update momentum coefficient
    update(m)
\end{lstlisting}
\begin{lstlisting}[language=python]
def hybrid_loss(o, t, c1, c2, r):
    # compute forward / backward pseudo-dynamic signals
    f_pd, b_pd = pseudo(o, t), pseudo(t, o)
    # compute the positive mask
    D = dist(c1, c2)
    A = (D <= r).detach()
    # compute static and dynamic similarities
    static = (torch.bmm(o.T, t) * A).sum()
    dynamic = (torch.bmm(f_op.T, b_op) * A).sum()
    # compute hybrid loss
    loss = - (static + dynamic) / (A.sum() + 1e-6)
    return loss
\end{lstlisting}
\end{algorithm}

\subsection{Hybrid Visual Correspondence Learning}\label{sec:3.2}
In recent years, self-supervised VOS has witnessed significant advancements, driven by the incorporation of various leading technologies such as \textit{photometric reconstruction}~\cite{vondrick2018tracking,lai2019self,lai2020mast,jeon2021mining,dense2021,wang2021contrastive}, \textit{cycle consistency tracking}~\cite{wang2019learning,wang2019unsupervised,jabri2020space,lu2020learning,zhao2021modelling,li2022liir}, and \textit{contrastive matching}~\cite{jeon2021mining,dense2021,xu2021rethinking,son2022contrastive}.
These methods share three key similarities: 1) They employ video data for training. 2) Their primary objective centers around spatio-temporal correspondence. 3) The affinity matrix serves as an intermediate variable for reconstruction. 

\subsubsection{Self-Supervised VOS in Videos} \label{sec:3.2.1}
Given the reference and query frames $\bm{I}_{ref}$, $\bm{I}_{query}$, we formulaically represent the affinity matrix $\bm{S}$ as follows:
\begin{equation}
    \begin{aligned}\label{eq:aff}
    \bm{S}_{ref}^{query} = \texttt{softmax}_{row}(\bm{I}_{ref} \bm{I}_{query}^{\top})\in [0,1]^{hw \times hw},
    \end{aligned}
\end{equation}
where $\texttt{softmax}_{row}$ represent row-wise \texttt{softmax}. In this manner, the three self-supervised strategies (\textit{photometric reconstruction} $\mathcal{L}_{pr}$, \textit{cycle consistency} $\mathcal{L}_{cc}$, and \textit{contrastive matching} $\mathcal{L}_{cm}$) can be expressed as follows:
\begin{equation}
    \begin{aligned}\label{eq:loss}
    \mathcal{L}_{pr} &= ||\bm{I}_{query} - {\bm{S}_{ref}^{query}}^{\top} \bm{I}_{ref}||^2, \\
    \mathcal{L}_{cc} &= ||\bm{S}_{ref}^{query} \bm{S}_{query}^{ref} - \mathbbm{1}||^2, \\
    \mathcal{L}_{cm} &= -\texttt{log}\frac{\texttt{exp}(\bm{S}(p,p^+))}{\sum_{p^-} \texttt{exp}(\bm{S}(p,p^-))},
    \end{aligned}
\end{equation}
where $\mathbbm{1}$ is an identity matrix with the appropriate size. $(p,p^+)$ indicate positive (matched) pixel pairs, and $(p,p^-)$ denote negative (unrelated) pixel pairs. It becomes apparent that the current approaches heavily depend on the affinity matrix $\bm{S}$. However, the acquisition of $\bm{S}$ is prone to interference from data noise~\cite{son2022contrastive}. 

If accurate correspondence between pixels can be acquired, the above affinity matrix-based methods could be largely improved.
While existing methods primarily focus on learning correspondence between video frames, we propose to approach this in a different way. 
Specifically, we randomly crop two overlapped views from the same image, simulating a pair of consecutive frames within a video sequence. In this case, the overlapped regions are considered as positive samples to learn a pseudo-dynamic correspondence. This new pseudo-dynamic correspondence learning approach will be elaborated on below.

\begin{figure}[t]
    \begin{center}
        \includegraphics[width=1.0\linewidth]{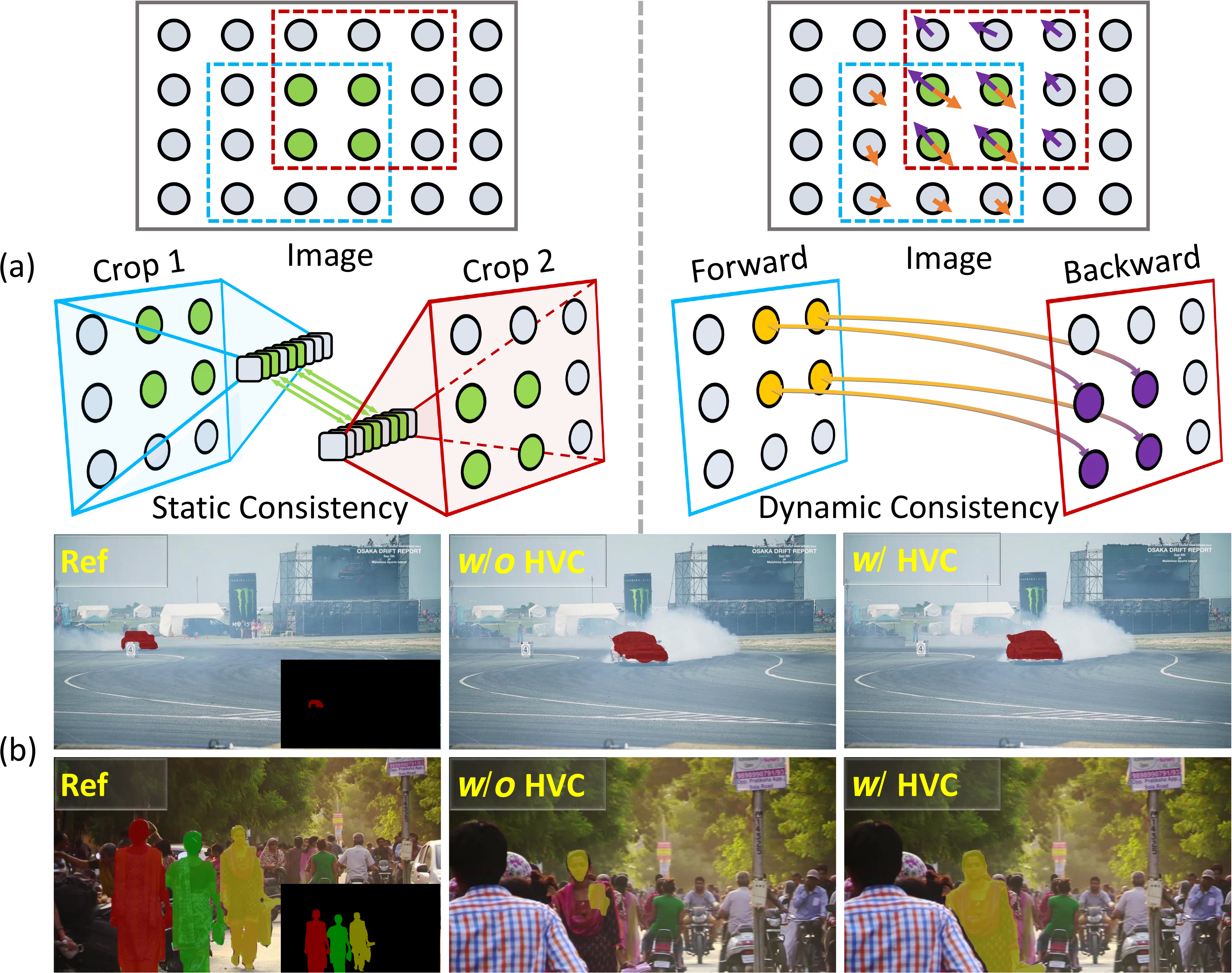}
        \put(-145.00,93.50){\scriptsize Eq.~\eqref{eq:2}}
        \put(-19.00,93.50){\scriptsize Eq.~\eqref{eq:3}}
    \end{center}
    \vspace{-6pt}
    \caption{(a) Illustration of \textbf{\textit{static} and \textit{dynamic} visual correspondence}. (b) Visualization of mask propagation.}\label{fig:arch}
\end{figure}

\subsubsection{Hybrid Visual Correspondence in Static Images} \label{sec:3.2.2}
Indeed, static images inherently lack realistic dynamic information. However, it is feasible to generate pseudo-dynamic signals by altering the viewpoint through cropping mechanisms is feasible~\cite{ye2021motion,liu2022content}. In this section, we comprehensively describe the process of learning static-dynamic visual correspondence utilizing image data.

\noindent\textbf{Pseudo-Dynamic Signal Generation.}
Existing video-based self-supervised VOS approaches have focused on learning the affinity matrix in video data due to its crucial role in ensuring static-dynamic consistency for VOS tasks. Our baseline approach constructed from Eq.~\eqref{eq:2} has a potential problem: the distance metric of the feature map only provides static consistency but not dynamic consistency.

To address the above issues, we introduce a hybrid visual correspondence approach, as depicted in Fig.~\ref{fig:arch}(a). Specifically, given the projection features $\bm{F}_1$ and $ \bm{F}_2$, we construct Eq.~\eqref{eq:2} to provide static consistency constraint. 
To model dynamic consistency, we propose a pseudo-dynamic signal generation method, which involves fake video frames by cropping image patches followed by pseudo-dynamic signal estimation (similar to optical flow) between these cropped image views.
Pseudo-dynamic signals are not direct optical flow estimations and guide the network in focusing on areas depicting motion between the pseudo-frames.
The estimation module $\mathcal{F}_{\rm{pseudo}}$ consists of two consecutive convolutional layers with \texttt{BN} and \texttt{ReLU}.
Particularly, our pseudo-dynamic signal generation method is seamlessly integrated into the entire framework without the need for any data annotation or conventional supervision. Together with Eq.~\eqref{eq:2}, the loss function is defined as:
\begin{equation}
    \begin{aligned}\label{eq:3}
        \mathcal{L}_{\rm{hybrid}} = -(\frac{<\bm{F}_1, \bm{F}_2>}{\begingroup
        \color{ggray}
        \underbrace{\color{black}\textstyle||\bm{F}_1||_{2}~||\bm{F}_2||_{2}}_{\substack{\text{ \footnotesize {\color{myred}\textit{static}} visual}\\ \text{ \footnotesize  correspondence}}} \endgroup}
        +
        \frac{<\bm{M}_1, \bm{M}_2>}{\begingroup
        \color{ggray}
        \underbrace{\color{black}\textstyle||\bm{M}_1||_{2}~||\bm{M}_2||_{2}}_{\substack{\text{ \footnotesize {\color{mygreen}\textit{dynamic}} visual} \\ \text{\footnotesize correspondence}}}\endgroup}), \\
        \bm{M}_1,~~\bm{M}_2 = \mathcal{F}_{\rm{pseudo}}(\bm{F}_1, \bm{F}_2),~~\mathcal{F}_{\rm{pseudo}}(\bm{F}_2, \bm{F}_1),
    \end{aligned}
\end{equation}
where $\bm{M}_1,\bm{M}_2\in\mathbb{R}^{{2}\times{H}\times{W}}$ indicate the forward and backward pseudo-dynamic signals between $\bm{F}_1$ and $\bm{F}_2$, respectively.
Despite the absence of explicit optical flow computation, our HVC effectively leverages the spatial transformations between overlapping regions to simulate motion.
The visual results are detailed in \S\ref{sec:dynamic_flow}. Our objective is to maximize the similarity of static and dynamic features.

\noindent\textbf{Positive Region Selection.}
Here, we will describe how we construct the positive pairs of feature maps ($\bm{F}_{\cdot}$ and  $\bm{M}_{\cdot}$).
As described in Eq.~\eqref{eq:2}, the sizes of two crop views in a given image differ, leading to shifts in feature positions after resizing.
We randomly crop the image to obtain two views and record their original coordinates (top-left corner and bottom-right corner). These cropped views and their respective coordinates are the resized and transformed following a standard protocol~\cite{xie2021propagate}.
Based on the spatial positions of the two cropped views in the original image, we generate the pixel coordinates of views after transformation as:
\begin{equation}
    \begin{aligned}\label{eq:8}
    \bm{x}^o_{1},\bm{y}^o_{1} = \mathcal{W}(\bm{x}_1, \bm{y}_1), \\
    \bm{x}^o_{2},\bm{y}^o_{2} = \mathcal{W}(\bm{x}_2, \bm{y}_2),
    \end{aligned}
\end{equation}
where $\mathcal{W}(\cdot, \cdot)$ indicates warping operation, and $\bm{x}^{o}_{\cdot},\bm{y}^{o}_{\cdot}\in\mathbb{R}^{{H}\times{W}}$ represent the transformed horizontal and vertical pixel coordinates, respectively.

Since the original coordinates of these views are known, we transform the projected features back to the original image space using the warping operation in Eq.~\eqref{eq:8}.
Subsequently, we can calculate the Euclidean distance $\mathcal{D}$ between the two coordinate sets of feature maps in the original image space (see Fig.~\ref{fig:ivr}). This distance is used to gauge feature similarities, contributing to the model’s ability to establish visual correspondence. To define the spatial neighborhood of the local feature space, we set a positive radius $r$ for the feature vectors that correspond to the identical cropped region (the overlapping region of the two crop views). Last, the positive region selection mask $\bm{A}\in\mathbb{R}^{{HW}\times{HW}}$ for the two crop views can be represented as:
\begin{equation}
    \begin{aligned}\label{eq:4}
       \bm{A}(i,j)=
       \begin{cases}
        1,  & {\rm{if}}~~\mathcal{D}(i,j)\leq{r}\\
        0,  & \rm{otherwise}
       \end{cases},
    \end{aligned}
\end{equation}
where $i,j$ represent the pixel coordinates from both crop views, and we set the positive radius $r$ to 0.1 (see \S\ref{sec:abs} for details). 
Unlike the affinity matrix-based approaches, we adopt a more direct optimization objective. To learn accurate visual correspondence from static images, we design the self-supervised signal (\ie, $\bm{A}$), guided by the known coordinates of the two image-cropped views. Thus, in conjunction with Eq.~\eqref{eq:4}, the static and dynamic visual correspondence methods described in Eq.~\eqref{eq:3} can be further expressed as:
\begin{equation}
    \begin{aligned}\label{eq:5}
    \bm{T}_{\rm{static}} = \frac{<\bm{F}_1, \bm{F}_2>}{||\bm{F}_1||_{2}~||\bm{F}_2||_{2}} \odot \bm{A} \in\mathbb{R}^{{HW}\times{HW}}, \\
    \bm{T}_{\rm{dynamic}} = \frac{<\bm{M}_1, \bm{M}_2>}{||\bm{M}_1||_{2}~||\bm{M}_2||_{2}} \odot \bm{A} \in\mathbb{R}^{{HW}\times{HW}},
    \end{aligned}
\end{equation}
where $\odot$ denotes the Hadamard product. We ultimately aim to maximize the similarity matrices $\bm{T}_{\rm{static}}$ and $\bm{T}_{\rm{dynamic}}$.

\begin{figure}[t]
    \begin{center}
        \includegraphics[width=1\linewidth]{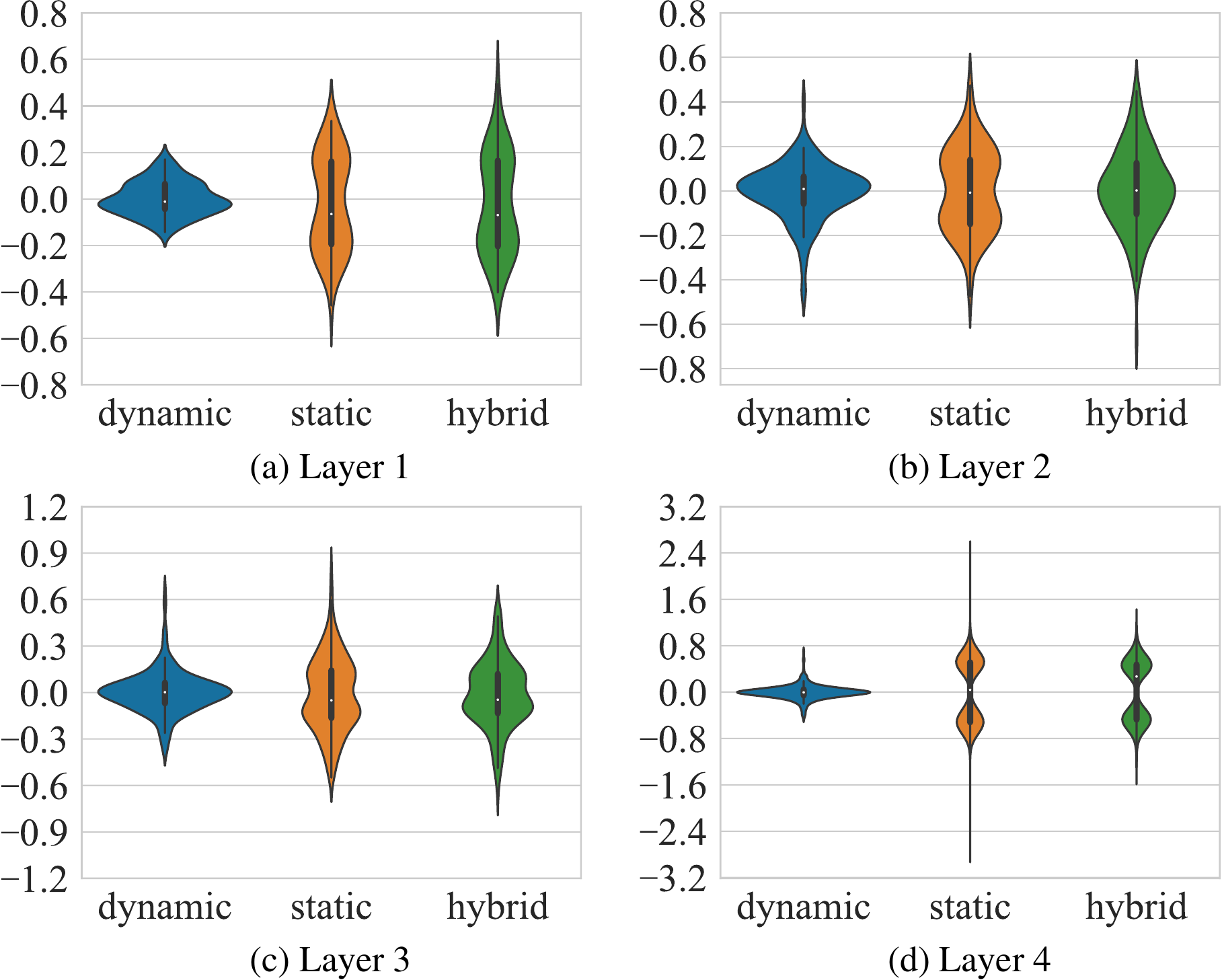}
    \end{center}
    \vspace{-6pt}
    \caption{Statistics of encoder weights (\S\ref{sec:3.2}) at each layer of ResNet-18. The weights learned by \textbf{dynamic}, \textbf{static} and \textbf{hybrid} approaches are displayed from left to right in each violin plot.}\label{fig:violin}
    \vspace{-6pt}
\end{figure}

\noindent\textbf{Hybrid Loss for Visual Correspondence Learning.}
Learning visual correspondences from static images and applying them to dense video segmentation downstream tasks is not trivial. Dynamic signals between video frames are simulated by generating pseudo-dynamic signals using image-cropped views. This approach allows us to construct static and dynamic feature similarity in images, complementing our proposed model's static and dynamic consistency. By combining Eq.~\eqref{eq:5}, we rewrite Eq.~\eqref{eq:3} to fit into hybrid loss:
\begin{equation}
    \begin{aligned}\label{eq:6}
    &\mathcal{L}_{\rm{static}}=-\frac{\sum_{i=1}^{HW}\sum_{j=1}^{HW}\bm{T}_{\rm{static}}(i,j)}{\sum_{i=1}^{HW}\sum_{j=1}^{HW}\bm{A}(i,j)}, \\
    &\mathcal{L}_{\rm{dynamic}}=-\frac{\sum_{i=1}^{HW}\sum_{j=1}^{HW}\bm{T}_{\rm{dynamic}}(i,j)}{\sum_{i=1}^{HW}\sum_{j=1}^{HW}\bm{A}(i,j)}, \\
    &\mathcal{L}_{\rm{hybrid}}=\mathcal{L}_{\rm{static}}+\alpha\mathcal{L}_{\rm{dynamic}},
    \end{aligned}
\end{equation}
where $\alpha$ represents the weight factor of the dynamic loss $\mathcal{L}_{\rm{dynamic}}$ (set to 1 by default, please refer to \S\ref{sec:abs} for details).
Using $\mathcal{L}_{\rm{hybrid}}$ as the loss function, we establish the hybrid visual correspondence framework by employing Eq.~\eqref{eq:1}. Examples of mask propagation are shown in Fig.~\ref{fig:arch}(b). 

To facilitate a more intuitive comparison of the dissimilarities in the learned feature representations of these approaches (\ie, static, dynamic, and hybrid), we extract and visualize the feature weights of each encoder ($\mathcal{L}_{\rm{staitc}}$, $\mathcal{L}_{\rm{dynamic}}$, and $\mathcal{L}_{\rm{hybrid}}$).
As depicted in Fig.~\ref{fig:violin}, the weights of dynamic correspondence learning appear more compact than the static ones. This compactness leads to a reduction in discrepancies among features obtained through the $\mathcal{L}_{\rm{dynamic}}$ projection. A narrow distribution, as observed for the dynamic correspondence weights, indicates a higher degree of agreement among the features after the $\mathcal{L}_{\rm{dynamic}}$ projection. On the other hand, the wider distribution of the static weights before the integration of dynamic correspondence learning implies a broader range of values and, subsequently, a more diverse set of feature representations. Upon integrating the hybrid correspondence, informed by both $\mathcal{L}_{\rm{static}}$ and $\mathcal{L}_{\rm{dynamic}}$, we observe a more integrated weight distribution that leverages the strengths of both correspondence types. The resulting feature representations are not only discriminative but also adept at generalizing across variations in motion and appearance in video sequences.

\begin{table*}[t]
    \caption{\textbf{Quantitative results for video object segmentation} (\S\ref{sec:vos}) on DAVIS$_{17}$~\cite{pont2017} \texttt{val-set}. (\textit{\textbf{\#}the number of images, \textbf{\#}videos duration in hours}) and (\textit{\textbf{\#}image-level annotations, \textbf{\#}pixel-level annotations}) denote the dataset sizes for the self-supervised and supervised settings, respectively. For evaluation metrics, $\cdot_{m}$ and $\cdot_{r}$ are the mean and recall ones. $\mathcal{J}\&\mathcal{F}_{m}$ indicates the average of $\mathcal{J}_{m}$ and $\mathcal{F}_{m}$. HVC-YT indicates that our model is pre-trained on YouTube-VOS$_{18}$~\cite{xu2018youtube} \texttt{train-set}, the same as in Tables~\ref{table:youtube}, \ref{table:davis16}, \ref{table:vost}, \ref{table:vip_jhmdb}, \ref{table:efficiency}, \ref{table:davis-17test} and \ref{table:ytb-2019}.}\label{table:davis17}
    \centering
        \resizebox{\textwidth}{!}{
        \setlength\tabcolsep{4.5pt}
	\renewcommand\arraystretch{1.0}
	\begin{tabular}{ccccccccccc}
            \hline\thickhline
		Method &Publication & Backbone & Supervised & Traning Dataset &Size & $\mathcal{J}\&\mathcal{F}_{m}$ $\uparrow$ & $\mathcal{J}_{m}$ $\uparrow$ & $\mathcal{J}_{r}$ $\uparrow$  &  $\mathcal{F}_{m}$ $\uparrow$  & $\mathcal{F}_{r}$ $\uparrow$  \\ \hline \hline
		Colorization~\cite{vondrick2018tracking} &ECCV-2018 & ResNet-18 & \xmark & Kinetics &(~-~, 800 hours)
		& 34.0 & 34.6 & 34.1 & 32.7 & 26.8 \\
		TimeCycle~\cite{wang2019learningcorr} &CVPR-2019 & ResNet-50 & \xmark & VLOG &(~-~, 344 hours)
		& 48.7 & 46.4 & 50.0 & 50.0 & 48.0 \\
		CorrFlow$^{\star}$~\cite{lai2019self} &BMVC-2019 & ResNet-18 & \xmark & OxUvA &(~-~, 14 hours)
		& 50.3 & 48.4 & 53.2 & 52.2 & 56.0 \\
		UVC~\cite{li2019joint} &NeurIPS-2019 & ResNet-50 & \xmark & C + Kinetics &(118K, 800 hours)
		& 60.9 & 59.3 & 68.8 & 62.7 & 70.9 \\
		MuG~\cite{lu2020learning} &CVPR-2020 & ResNet-18 & \xmark & OxUvA &(~-~, 14 hours)
		& 54.3 & 52.6 & 57.4 & 56.1 & 58.1  \\
		MAST~\cite{lai2020mast} &CVPR-2020 & ResNet-18 & \xmark & YT &(~-~, 5.58 hours)
		& 65.5 & 63.3 & 73.2 & 67.6 & 77.7  \\
		CRW~\cite{jabri2020space} &NeurIPS-2020 & ResNet-18 & \xmark & Kinetics &(~-~, 800 hours)
		& 67.6 & 64.8 & 76.1 & 70.2 & 82.1 \\
		ContrastCorr~\cite{wang2021contrastive} &AAAI-2021 & ResNet-18 & \xmark & C + TN &(118K, 300 hours)
		& 63.0 & 60.5 & - & 65.5 & - \\
		VFS~\cite{xu2021rethinking} &ICCV-2021 & ResNet-18 & \xmark & Kinetics &(~-~, 800 hours)
		& 66.7 & 64.0 & - & 69.4 & - \\
		JSTG~\cite{zhao2021modelling} &ICCV-2021 & ResNet-18 & \xmark & Kinetics &(~-~, 800 hours)
		& 68.7 & 65.8 & 77.7  & 71.6 & 84.3 \\
            CLTC$^{\star}$~\cite{jeon2021mining} &CVPR-2021 & ResNet-18 & \xmark & YT &(~-~, 5.58 hours) & 70.3 & 67.9 & 78.2 & 72.6 & 83.7 \\
            DUL~\cite{dense2021} &NeurIPS-2021 & ResNet-18 & \xmark & YT &(~-~, 5.58 hours) & 69.3 & 67.1 & 81.2 & 71.6 & 84.9 \\
		SFC~\cite{hu2022sfc} &ECCV-2022 & ResNet-18 & \xmark & YT &(~-~, 5.58 hours)
		& 71.2 & 68.3 & - & 74.0 & - \\
  		SCC~\cite{son2022contrastive} &CVPR-2022 & ResNet-18 & \xmark & YT &(~-~, 5.58 hours)
		& 70.5 & 67.4 & 78.8 & 73.6 & 84.6 \\
  		LIIR$^{\star\dag}$~\cite{li2022liir} &CVPR-2022 & ResNet-18 & \xmark & YT &(~-~, 5.58 hours)
		& 72.1 & 69.7 & 81.4 & 74.5 & 85.9 \\
  		\textbf{HVC-YT~(ours)} &~-~ & ResNet-18 & \xmark & YT &(~-~, 5.58 hours) & \textbf{73.1} & \textbf{70.3} & 83.0 & \textbf{75.9} & 86.8 \\  
            \hdashline
		PixPro~\cite{xie2021propagate} &CVPR-2021 & ResNet-50 & \xmark & IN &(1.28M, ~-~)
		& 58.9 & 57.9 & 69.9  & 59.8 & 67.2 \\
            DINO~\cite{caron2021emerging} &ICCV-2021 & ViT-B/8 & \xmark & IN &(1.28M, ~-~) & 71.4 & 67.9 & - & 74.9 & - \\
            ODIN~\cite{ODIN} &ECCV-2022 & ResNet-50 & \xmark & IN &(1.28M, ~-~) & 54.1 & 54.3 & - & 53.9 & - \\
            CLIP-S$^4$~\cite{CLIP-S4} &CVPR-2023 & ResNet-50 & \xmark & IN &(1.28M, ~-~) & 54.6 & 52.3 & - & 56.8 & - \\
  		CrOC~\cite{CrOC} &CVPR-2023 & ViT-S/16 & \xmark & C &(118K, ~-~) & 58.4 & 56.5 & - & 60.2 & - \\
  		\rowcolor{rowgray} \textbf{HVC~(ours)} &~-~ & \textbf{ResNet-18} & \xmark & C &(118K, ~-~) & \textbf{73.1} & \textbf{70.3} & \textbf{83.1} & 75.8 & \textbf{86.9} \\  
		\hline
		\color{mygray1}ResNet~\cite{he2016deep} &\color{mygray1}CVPR-2016 & \color{mygray1}ResNet-18 & \color{mygray1}\cmark & \color{mygray1}IN &\color{mygray1}(1.28M, ~-~)
		& \color{mygray1}62.9 & \color{mygray1}60.6 & \color{mygray1}69.9 & \color{mygray1}65.2 & \color{mygray1}73.8 \\
		\color{mygray1}OSVOS~\cite{caelles2017one} &\color{mygray1}CVPR-2017 & \color{mygray1}VGG-16 & \color{mygray1}\cmark & \color{mygray1}IN+D &\color{mygray1}(1.28M, 10K)
		& \color{mygray1}60.3 & \color{mygray1}56.6 & \color{mygray1}63.8 & \color{mygray1}63.9 & \color{mygray1}73.8 \\
		\color{mygray1}STM~\cite{oh2019video} &\color{mygray1}CVPR-2019 & \color{mygray1}ResNet-50 & \color{mygray1}\cmark & \color{mygray1}IN + D + YT &\color{mygray1}(1.28M, 164K)
		& \color{mygray1}81.8 & \color{mygray1}79.2 & \color{mygray1}88.7 & \color{mygray1}84.3 & \color{mygray1}91.8 \\
            \color{mygray1}XMem~\cite{cheng2022xmem} &\color{mygray1}ECCV-2022 & \color{mygray1}ResNet-50 & \color{mygray1}\cmark & \color{mygray1}IN + D + YT &\color{mygray1}(1.28M, 164K)
		& \color{mygray1}86.2 & \color{mygray1}82.9 & \color{mygray1}- & \color{mygray1}89.5 & \color{mygray1}- \\ \hline
		\end{tabular}
	}
    \leftline{~~\footnotesize{$^{\star}$: 2$\times$ resolution. $^\dag$: multi-step training. IN: ImageNet~\cite{deng2009imagenet}. C: COCO~\cite{lin2014microsoft}. D: DAVIS~\cite{pont2017}. YT: YouTube-VOS~\cite{xu2018youtube}. TN: TrackingNet~\cite{muller2018trackingnet}.}}
\end{table*}

\begin{figure*}[t]
    \begin{center}
        \includegraphics[width=1.0\linewidth]{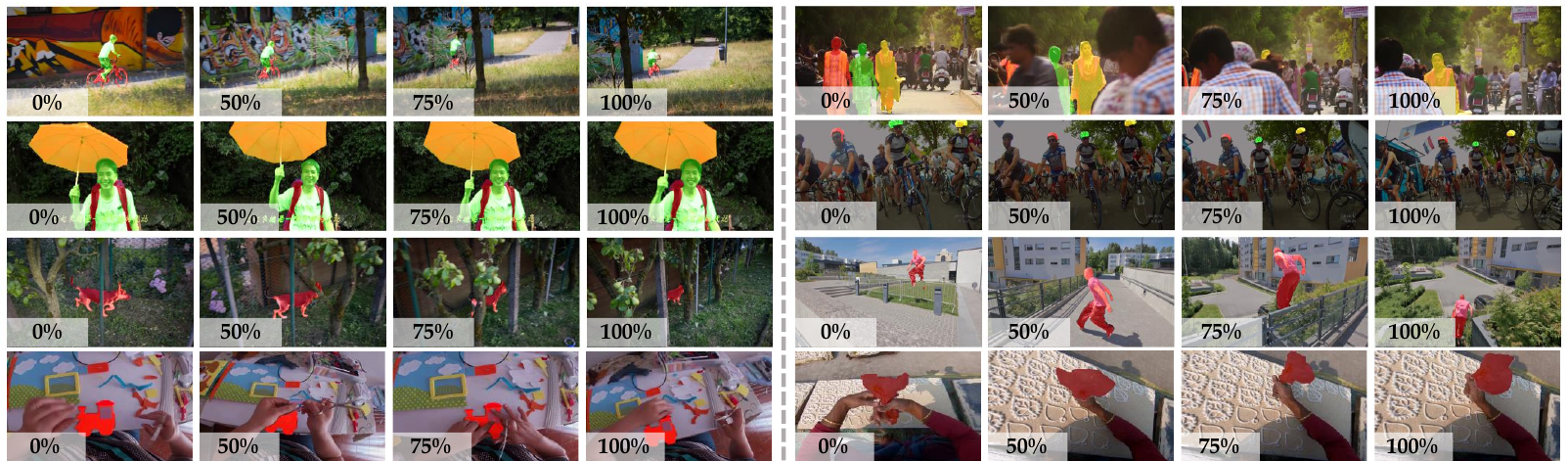}
    \end{center}
    \caption{\textbf{Qualitative results for video object segmentation} (\S\ref{sec:vos}) on DAVIS$_{17}$~\cite{pont2017} \texttt{val-set} (1st row), YouTube-VOS$_{18}$~\cite{xu2018youtube} \texttt{val-set} (2nd row), DAVIS$_{16}$~\cite{perazzi2016benchmark} \texttt{val-set} (3rd row), and VOST~\cite{tokmakov2023breaking} \texttt{val-set} (4th row). `\%' indicates progress status. Zoomed-in view for best.}\label{fig:results_vis}
\end{figure*}

\subsection{Implementation Details}\label{sec:4} \label{sec:3.3}
\noindent\textbf{Architecture.} Similar to previous studies~\cite{jabri2020space,lai2020mast,son2022contrastive,li2022liir,dense2021,xu2021rethinking}, we use ResNet-18~\cite{he2016deep} as the backbone to extract features. We remove the strides in the last two residual blocks (\texttt{res3} and \texttt{res4}) like~\cite{wang2019learningcorr,li2019joint,lu2020learning,hu2022sfc}. The space projection module (see \S\ref{sec:3.1} for details) constructs the projector head and predictor head for feature embeddings on backbone output. We perform the $l_{2}$ normalization to obtain feature embeddings.

\noindent\textbf{Training.} We utilize COCO~\cite{lin2014microsoft} as the static training dataset and train our hybrid visual correspondence model from scratch using a single NVIDIA A100 GPU.
During training, the label-free raw images are randomly sampled. We create two randomly generated crops using only spatial augmentation, and resize all image crops to 256$\times$256, following~\cite{jabri2020space,lai2020mast,son2022contrastive,li2022liir,dense2021,xu2021rethinking}. The entire model is implemented based on the PyTorch framework~\cite{paszke2019pytorch}, with a mini-batch size of 128 for training over 20 epochs. The Adam optimizer~\cite{kingma2014adam} is adopted, with an initial learning rate of $10^{-3}$ and a weight decay of 0. At training time, the online encoder is optimized by gradients, while the target encoder updates its parameters by the moving average in each iteration. Following ~\cite{he2020momentum}, the default value starts at 0.99 and increases gradually to 1.

\noindent\textbf{Inference.} Our HVC directly provides learned feature representations without any fine-tuning. Similar to~\cite{jabri2020space,dense2021,caron2021emerging}, we adopt a versatile label propagation solution to evaluate VOS. Specifically, the first frame labels are provided for a video, and we compute the feature embedding similarity between the current frame and the previous frames (to provide first frame labels and previous predictions) to propagate labels to remaining video frames. Analogous to \cite{hu2022sfc}, we employ MoCo~\cite{he2020momentum} as a baseline model to acquire rich semantic representations. To verity its effectiveness, HVC is evaluated on various challenging VOS datasets.

\begin{table}
    \caption{\textbf{Quantitative results for video object segmentation} (\S\ref{sec:vos}) on YouTube-VOS$_{18}$~\cite{xu2018youtube} \texttt{val-set}.}\label{table:youtube}
    \centering
	\resizebox{\columnwidth}{!}{
	\setlength\tabcolsep{5.5pt}
	\renewcommand\arraystretch{1.05}
	\begin{tabular}{ccccccc}
            \hline\thickhline
            & & & \multicolumn{2}{c}{Seen} & \multicolumn{2}{c}{Unseen} \\ \cline{4-7}
            \multirow{-2}{*}{Method} & \multirow{-2}{*}{Sup.} & \multirow{-2}{*}{Mean} & $\mathcal{J}_{m}$ $\uparrow$ & $\mathcal{F}_{m}$ $\uparrow$ & $\mathcal{J}_{m}$ $\uparrow$ & $\mathcal{F}_{m}$ $\uparrow$\\ \hline\hline
		Colorization~\cite{vondrick2018tracking} & \xmark & 38.9 & 43.1 & 38.6 & 36.6 & 37.4 \\
		CorrFlow$^{\star}$~\cite{lai2019self} & \xmark & 46.6 & 50.6& 46.6 & 43.8 & 45.6 \\
		MAST~\cite{lai2020mast} & \xmark & 64.2 & 63.9 & 64.9 & 60.3 & 67.7 \\
		CLTC$^{\star}$~\cite{jeon2021mining} & \xmark & 67.3 & 66.2 & 67.9 & 63.2& 71.7 \\
		LIIR$^{\star\dag}$~\cite{li2022liir} & \xmark & 69.3 & 67.9 & 69.7 & 65.7 & 73.8 \\
		CRW~\cite{jabri2020space} & \xmark & 69.9 & 68.7 & 70.2 & 65.4 & 75.2 \\
		DUL~\cite{dense2021} & \xmark & 69.9 & 69.6 & 71.3 & 65.0 & 73.5 \\
            \hdashline
		\textbf{HVC-YT~(ours)} & \xmark & 71.6 & 70.0 & 72.1 & 67.3 & 77.0 \\  
  		\rowcolor{rowgray} \textbf{HVC~(ours)} & \xmark & \textbf{71.9} & \textbf{70.3} & \textbf{72.4} & \textbf{67.6} & \textbf{77.2} \\  
		\hline
		\color{mygray1}OSVOS~\cite{caelles2017one} & \color{mygray1}\cmark & \color{mygray1}58.8 & \color{mygray1}59.8&~\color{mygray1}60.5 & \color{mygray1}54.2 & \color{mygray1}60.7 \\
		\color{mygray1}PreMVOS~\cite{luiten2018premvos} & \color{mygray1}\cmark & \color{mygray1}66.9 & \color{mygray1}71.4&~\color{mygray1}75.9 & \color{mygray1}56.5 & \color{mygray1}63.7 \\
		\color{mygray1}STM~\cite{oh2019video} & \color{mygray1}\cmark & \color{mygray1}79.4 & \color{mygray1}79.7 & \color{mygray1}84.2 & \color{mygray1}72.8&\color{mygray1} 80.9
		\\ \hline
	\end{tabular}
	}
        \leftline{~~\footnotesize{$^{\star}$: 2$\times$ resolution. $^\dag$: multi-step training.}}
\end{table}

\begin{table}
    \caption{\textbf{Quantitative results for video object segmentation} (\S\ref{sec:vos}) on DAVIS$_{16}$~\cite{perazzi2016benchmark} \texttt{val-set}.}\label{table:davis16}
    \centering
        \resizebox{\columnwidth}{!}{
	\setlength\tabcolsep{4.5pt}
	\renewcommand\arraystretch{1.05}
	\begin{tabular}{ccccccc}
		\hline\thickhline
            Method & Sup. & $\mathcal{J}\&\mathcal{F}_{m}$ $\uparrow$ & $\mathcal{J}_{m}$ $\uparrow$ & $\mathcal{J}_{r}$ $\uparrow$ & $\mathcal{F}_{m}$ $\uparrow$ & $\mathcal{F}_{r}$ $\uparrow$ \\ \hline\hline
		Colorization~\cite{vondrick2018tracking} & \xmark & 34.9 & 38.9 & 37.1 & 30.0 & 21.7 \\
		CorrFlow$^{\star}$~\cite{lai2019self} & \xmark & 48.0 & 47.1 & 51.3 & 49.9 & 52.4 \\
		TimeCycle~\cite{wang2019learningcorr} & \xmark & 53.5 & 55.8 & 64.9 & 51.1 & 51.6 \\
		MAST~\cite{lai2020mast} & \xmark & 68.8 & 69.3 & 82.7 & 68.3 & 78.9 \\
		MPFM~\cite{li2022pixels} & \xmark & 74.3 & 75.2 & 89.5 & 73.3 & 82.0 \\
		SPGO$^\ddag$~\cite{ponimatkin2022simple} & \xmark & 78.8 & \textbf{80.2} & - & 77.5 & -  \\
            \hdashline
		\textbf{HVC-YT~(ours)} & \xmark & \textbf{80.1} & 79.1 & 91.6 & \textbf{81.0} & 88.6 \\ 
            \rowcolor{rowgray} \textbf{HVC~(ours)} & \xmark & \textbf{80.1} & 79.3 & \textbf{91.9} & 80.9 & \textbf{89.0} \\ 
		\hline
		\color{mygray1}OSVOS~\cite{caelles2017one} & \color{mygray1}\cmark & \color{mygray1}80.2 & \color{mygray1}79.8 &~\color{mygray1}93.6 & \color{mygray1}80.6 & \color{mygray1}92.6 \\
  		\color{mygray1}SiamMask~\cite{wang2019fast} & \color{mygray1}\cmark & \color{mygray1}70.0 & \color{mygray1}71.7 &~\color{mygray1}86.8 & \color{mygray1}67.8 & \color{mygray1}79.8 \\
		\color{mygray1}STM~\cite{oh2019video} & \color{mygray1}\cmark & \color{mygray1}89.3 & \color{mygray1}88.7 & \color{mygray1}- & \color{mygray1}89.9 &\color{mygray1} -
		\\ \hline
	\end{tabular}
	}
        \leftline{~~\footnotesize{$^{\star}$: 2$\times$ resolution. $^\ddag$: using additional optical flow.}}
\end{table}

\section{Experiments}
Following established conventions~\cite{jabri2020space,dense2021,li2022liir,son2022contrastive}, the evaluation model requires inferring the remaining frames' pixel-level masks by providing the first frame mask of each video sequence. In \S\ref{sec:sota}, we utilize diverse video label propagation tasks to evaluate the learned representation of our model. The proposed critical components are extensively ablated in \S\ref{sec:abs}, and further analysis of HVC is conducted in \S\ref{sec:FurExp}.
To maintain experimental impartiality, we supply a version of our model, denoted as HVC-YT, pre-trained on the YouTube-VOS$_{18}$~\cite{xu2018youtube} training set. It should be noted that, unless specified, all primary results are derived from HVC trained exclusively on the static image dataset, \ie, COCO.

\subsection{Comparison with State-of-the-Art} \label{sec:sota}
In \S\ref{sec:vos}, we perform experimental comparisons of the proposed HVC with state-of-the-art VOS methods. Furthermore, we employ the proposed method to tackle two additional challenging video label propagation tasks: body part propagation in \S\ref{sec:bpp} and human pose tracking in \S\ref{sec:hpt}.

\begin{table}
    \caption{\textbf{Quantitative results for video object segmentation} (\S\ref{sec:vos}) on VOST~\cite{tokmakov2023breaking} \texttt{val-set}. $\mathcal{J}^{last}_{\cdot}$ is the last 25\% of the frames in a video.}\label{table:vost}
    \centering
        \resizebox{\columnwidth}{!}{
	\setlength\tabcolsep{6.5pt}
	\renewcommand\arraystretch{1.05}
	\begin{tabular}{cccccc}
		\hline\thickhline
            Method & Sup. & $\mathcal{J}_{m}$ $\uparrow$ & $\mathcal{J}_{r}$ $\uparrow$ & $\mathcal{J}^{last}_{m}$ $\uparrow$ & $\mathcal{J}^{last}_{r}$ $\uparrow$ \\ \hline\hline
		MoCo~\cite{he2020momentum} & \xmark & 18.1 & 18.1 & 10.2 & 8.8 \\
		PixPro~\cite{xie2021propagate} & \xmark & 16.8 & 16.8 & 10.0 & 9.5 \\
            DINO~\cite{caron2021emerging} & \xmark & 18.8 & 18.9 & 10.4 & 9.4 \\
            CRW~\cite{jabri2020space} & \xmark & 22.3 & 23.0 & 13.3 & 12.6 \\
            DUL~\cite{dense2021} & \xmark & 23.9 & 23.9 & 14.6 & 13.7 \\
            SFC~\cite{hu2022sfc} & \xmark & 20.6 & 21.0 & 11.8 & 11.1 \\
            \hdashline
		\textbf{HVC-YT~(ours)} & \xmark & 26.0 & \textbf{26.9} & 14.5 & 13.0 \\ 
            \rowcolor{rowgray} \textbf{HVC~(ours)} & \xmark & \textbf{26.3} & \textbf{26.9} & \textbf{15.3} & \textbf{14.0} \\ 
		\hline
  		\color{mygray1}HODOR-Img~\cite{athar2022hodor} & \color{mygray1}\cmark & \color{mygray1}24.2 &~\color{mygray1}- & \color{mygray1}13.9 & \color{mygray1}- \\
            \color{mygray1}XMem~\cite{cheng2022xmem} & \color{mygray1}\cmark & \color{mygray1}44.1 &~\color{mygray1}- & \color{mygray1}33.8 & \color{mygray1}- \\
		\color{mygray1}AOT~\cite{yang2021aot} & \color{mygray1}\cmark & \color{mygray1}48.7 & \color{mygray1}- & \color{mygray1}36.4 &\color{mygray1} -
		\\ \hline
	\end{tabular}
	}
\end{table}

\subsubsection{Results for Video Object Segmentation} \label{sec:vos}
\noindent\textbf{Dataset.} We evaluate performance on two publicly popular datasets: DAVIS$_{17}$~\cite{pont2017} and YouTube-VOS$_{18}$~\cite{xu2018youtube}, which contain 30 and 474 validation videos for performing the multi-object VOS task, respectively.
We also conduct a comparison experiment for single-object VOS on DAVIS$_{16}$~\cite{perazzi2016benchmark}. In addition, we include the challenging VOST~\cite{tokmakov2023breaking} benchmark, which comprises a validation set of 70 videos characterized by significant object transformation (breaking, occlusion, deformation, fast object motion, \etc) scenarios.

\noindent\textbf{Metric.} For quantitative comparison, we report the performance scores employing two official evaluation metrics~\cite{perazzi2016benchmark}, \ie, region similarity $\mathcal{J}$ and boundary accuracy $\mathcal{F}$. Given that YouTube-VOS consists of \textit{seen} and \textit{unseen} categories, we calculate the $\mathcal{J}$ and $\mathcal{F}$ scores separately for each category.

\noindent\textbf{Evaluation on DAVIS$_{17}$.} As shown in Table~\ref{table:davis17}, we compare our proposed method with state-of-the-art self-supervised methods. Our HVC scores \textbf{73.1\%} $\mathcal{J}\&\mathcal{F}_{m}$ and outperforms all other methods across all metrics. In contrast to LIIR~\cite{li2022liir}, which utilizes a multi-step training strategy, HVC reaches \textbf{1\%} $\mathcal{J}\&\mathcal{F}_{m}$ improvement with a more concise and efficient training approach (see Table~\ref{table:efficiency}). In addition, HVC delivers competitive performance compared to supervised methods.

\noindent\textbf{Evaluation on YouTube-VOS$_{18}$.} The results of HVC versus existing self-supervised methods on YouTube-VOS$_{18}$~\cite{xu2018youtube} \texttt{val-set} are shown in Table~\ref{table:youtube}. HVC achieves a new state-of-the-art and surpasses other self-supervised methods consistently. Compared to CRW~\cite{jabri2020space} and DUL~\cite{dense2021}, our method achieves a \textbf{2.0\%} higher mean score. Further, HVC is even superior to some well-known supervised methods~\cite{caelles2017one,luiten2018premvos}. The high scores obtained for both \textit{seen} and \textit{unseen} categories highlight the robustness and adaptability of our approach.

\noindent\textbf{Evaluation on DAVIS$_{16}$.} As illustrated in Table~\ref{table:davis16}, HVC outperforms all self-supervised methods in the single-object VOS setting. Compared to the state-of-the-art SPGO~\cite{ponimatkin2022simple}, which uses additional optical flow as auxiliary information, our approach improves by \textbf{1.3\%} in terms of $\mathcal{J}\&\mathcal{F}_{m}$. HVC particularly achieves a comparable performance to the renowned supervised method OSVOS~\cite{caelles2017one}.

\noindent\textbf{Evaluation on VOST.} HVC achieves leading performance among self-supervised approaches, scoring \textbf{15.3\%} $\mathcal{J}^{last}_{m}$, as detailed in Table~\ref{table:vost}.
Compared to DUL~\cite{dense2021}, our approach improves by \textbf{2.4\%}/\textbf{3.0\%} in terms of $\mathcal{J}_{m}$/$\mathcal{J}_{r}$. HVC demonstrates superior capabilities when confronting the demanding VOST~\cite{tokmakov2023breaking} benchmark (\eg, occlusion and deformation).

\noindent\textbf{Qualitative Results.}
Fig.~\ref{fig:results_vis} shows the video label propagation results from leveraging our learned visual correspondences. Our HVC has shown good segmentation results in confronting challenges such as occlusion, deformation, appearance changes, small objects, and fast motion.

\begin{figure}[t]
    \begin{center}
        \includegraphics[width=1\linewidth]{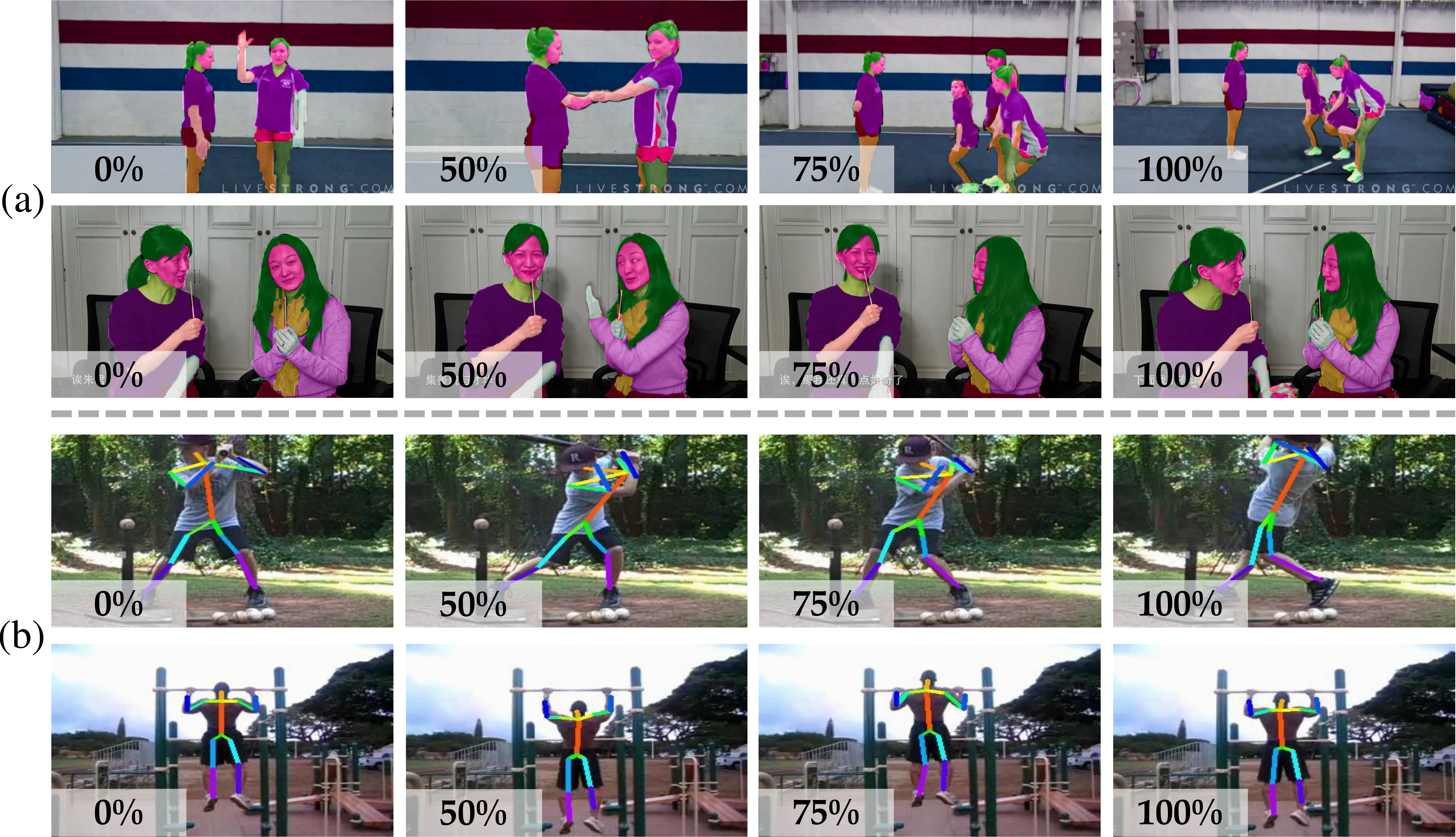}
    \end{center}
    \vspace{-6pt}
    \caption{\textbf{Qualitative results of the proposed method} for (a) \textbf{body part propagation} (\S\ref{sec:bpp}) on VIP~\cite{zhou2018adaptive} \texttt{val-set} and (b) \textbf{human pose tracking} (\S\ref{sec:hpt}) on JHMDB~\cite{jhuang2013towards} \texttt{val-set}. Zoomed-in view for best.}\label{fig:pose_tracking}
\end{figure}

\begin{table}[t]
    \caption{\textbf{Quantitative results for part propagation} (\S\ref{sec:bpp}) on VIP~\cite{zhou2018adaptive} \texttt{val-set} and \textbf{pose tracking} (\S\ref{sec:hpt}) on JHMDB~\cite{jhuang2013towards} \texttt{val-set}.}\label{table:vip_jhmdb}
    \centering
	\resizebox{\columnwidth}{!}{
	\setlength\tabcolsep{5.5pt}
	\renewcommand\arraystretch{1.0}
	\begin{tabular}{ccc||cc}
            \hline\thickhline
            & & \multicolumn{1}{c||}{VIP} & \multicolumn{2}{c}{JHMDB} \\ \cline{3-5}
            \multirow{-2}{*}{Method} & \multirow{-2}{*}{Sup.} & mIoU $\uparrow$ & PCK@0.1 $\uparrow$ & PCK@0.2 $\uparrow$\\ \hline\hline
		TimeCycle~\cite{wang2019learningcorr}  & \xmark & 28.9 & 57.3 & 78.1 \\
		UVC~\cite{li2019joint}                 & \xmark & 34.1 & 58.6 & 79.6 \\
            CRW~\cite{jabri2020space}              & \xmark & 38.6 & 59.3 & 80.3 \\
            VFS~\cite{xu2021rethinking}            & \xmark & 39.9 & 60.5 & 79.5 \\
		CLTC$^{\star}$~\cite{jeon2021mining}   & \xmark & 37.8 & 60.5 & 82.3 \\
            PixPro~\cite{xie2021propagate}         & \xmark & 29.6 & 57.8 & 80.8 \\
		LIIR$^{\star\dag}$~\cite{li2022liir}   & \xmark & 41.2 & 60.7 & 81.5 \\
            SCC~\cite{son2022contrastive}          & \xmark & 40.8 & \textbf{61.7} & 82.6 \\
            \hdashline
		\textbf{HVC-YT~(ours)} & \xmark & \textbf{45.1} & \textbf{61.7} & \textbf{82.8} \\ 
  		\rowcolor{rowgray} \textbf{HVC~(ours)} & \xmark & 44.6 & \textbf{61.7} & \textbf{82.8} \\ 
		\hline
		\color{mygray1}ResNet~\cite{he2016deep} & \color{mygray1}\cmark &\color{mygray1}31.9 & \color{mygray1}53.8 & \color{mygray1}74.6 \\
		\color{mygray1}TSN~\cite{song2017thin} & \color{mygray1}\cmark &\color{mygray1}- & \color{mygray1}68.7 & \color{mygray1}92.1 \\
		\color{mygray1}ATEN~\cite{zhou2018adaptive} & \color{mygray1}\cmark & \color{mygray1}37.9 & \color{mygray1}- &\color{mygray1}-
		\\ \hline
	\end{tabular}
	}
        \leftline{~~\footnotesize{$^{\star}$: 2$\times$ resolution. $^\dag$: multi-step training.}}
\end{table}

\subsubsection{Results for Body Part Propagation} \label{sec:bpp}
\noindent\textbf{Dataset.} We evaluate HVC on the Video Instance Parsing (VIP) dataset~\cite{zhou2018adaptive} for body part propagation. The \texttt{val-set} split of the benchmark contains 50 videos focusing on propagating 19 body parts, \eg, arms and legs. Hence, this task demands a higher level of precision in matching compared to video object segmentation. We follow the same settings as CRW~\cite{jabri2020space} and resize the video frames to $560 \times 560$.

\noindent\textbf{Metric.} To evaluate the performance of our model on the semantic-level propagation task, we use the standard metric provided by \cite{zhou2018adaptive}, \ie, mean intersection-over-union (mIoU).

\noindent\textbf{Evaluation on VIP.} Table~\ref{table:vip_jhmdb} (left) compares our model's performance with existing self-supervised methods. Our HVC achieves superior performance compared to LIIR~\cite{li2022liir} and SCC~\cite{son2022contrastive} by \textbf{3.4\%} and \textbf{3.8\%} in terms of mIoU, respectively. Moreover, HVC significantly improves over the fully supervised model ATEN~\cite{zhou2018adaptive} by \textbf{6.7\%}, which is a dedicated method designed for the VIP~\cite{zhou2018adaptive} dataset.

\noindent\textbf{Qualitative Results.} Fig.~\ref{fig:pose_tracking}(a) presents samples depicting the visual results of body part propagation. Our model adeptly propagates each part mask onto analogous instances.

\begin{table}[t]
    \caption{\textbf{Quantitative results of proposed key modules} (\S\ref{sec:abs}) on DAVIS$_{16, 17}$~\cite{perazzi2016benchmark,pont2017} \texttt{val-set}.}\label{table:ablation1}
    \centering
    \resizebox{\columnwidth}{!}{
    \setlength\tabcolsep{5pt}
    \renewcommand\arraystretch{1.1}
    \begin{tabular}{c|cc|c||c}
        \hline\thickhline
	& Static Visual & \multicolumn{1}{c|}{Dynamic Visual}& \multicolumn{1}{c||}{DAVIS$_{17}$} & \multicolumn{1}{c}{DAVIS$_{16}$} \\ \cline{4-5}
        \multirow{-2}{*}{\#}&\multirow{1}{*}{Correspondence} & Correspondence & $\mathcal{J}$\&$\mathcal{F}_m$ $\uparrow$  & $\mathcal{J}$\&$\mathcal{F}_m$ $\uparrow$ \\ \hline\hline
	1&   &    & 67.1 & 72.5 \\ \hline
        2&  \cmark &  & 70.6 \color{myred}{(\textbf{+3.5})}&  77.4 \color{myred}{(\textbf{+4.9})}\\
        3&  & \cmark & 69.9 \color{myred}{(\textbf{+2.8})}&  77.0 \color{myred}{(\textbf{+4.5})} \\
	4& \cmark & \cmark  & \textbf{73.1} \color{myred}{(\textbf{+6.0})} & \textbf{80.1} \color{myred}{(\textbf{+7.6})}\\
	\hline
    \end{tabular}
    }
\end{table}

\subsubsection{Results for Human Pose Tracking} \label{sec:hpt}
\noindent\textbf{Dataset.} We conduct the human pose tracking task using the JHMDB~\cite{jhuang2013towards} \texttt{val-set}, which contains 268 videos and entails detecting 15 human keypoints. Following the evaluation protocol of \cite{li2019joint,jabri2020space,son2022contrastive}, we evaluate our model with video frames resized to a resolution of $320\times320$.

\noindent\textbf{Metric.} We utilize the Possibility of Correct Keypoints (PCK) \cite{song2017thin} as the quantitative evaluation metric, which measures the percentage of keypoints that are in close proximity to the ground-truth keypoints across different thresholds.

\noindent\textbf{Evaluation on JHMDB.} We show the quantitative results of HVC against others in Table~\ref{table:vip_jhmdb} (right). HVC achieves consistent improvements over existing self-supervised methods on this challenging task that requires precise fine-grained matching. Remarkably, our model in PCK@0.1 and PCK@0.2 performs even better than the fully supervised baseline~\cite{he2016deep} (trained with labels) by \textbf{7.9\%} and \textbf{8.2\%}, respectively. Still, there is a gap between task-specific approaches (\eg, \cite{song2017thin}).

\noindent\textbf{Qualitative Results.} The pose tracking visualizations in Fig.~\ref{fig:pose_tracking}(b) affirms that HVC establishes the exact visual correspondence and effectively tracks human keypoints.

\subsection{Ablation Studies} \label{sec:abs}
Elaborated ablation studies are conducted on DAVIS~\cite{perazzi2016benchmark,pont2017} \texttt{val-set} to investigate our proposed approach.

\begin{table*}[t]
    \renewcommand\thetable{7}
    \caption{\textbf{Ablation studies} (\S\ref{sec:abs}) on DAVIS~\cite{perazzi2016benchmark,pont2017} \texttt{val-set}. Best results are marked in {\colorbox{mygray}{gray}}, representing the default settings of HVC. For baseline selection, ``Score A$\rightarrow${\color{mygreen}{\textbf{Score B}}}'' refers to the score of the original semantic model and the score after joining our method.}\label{tab:ablations}
    \vspace{-12pt}
    \subfloat[{Step between Images} \label{table:ablation2}]{
    \resizebox{0.209\textwidth}{!}{
    \setlength\tabcolsep{1.8pt}
    \begin{tabular}{c|c|c||c}
        \hline\thickhline
	&Step & \multicolumn{1}{c||}{DAVIS$_{17}$} & \multicolumn{1}{c}{DAVIS$_{16}$} \\ \cline{3-4}
        \multirow{-2}{*}{\#}&\multirow{-1}{*}{Size} & $\mathcal{J}$\&$\mathcal{F}_m$ $\uparrow$  & $\mathcal{J}$\&$\mathcal{F}_m$ $\uparrow$
        \\ \hline\hline
        \rowcolor{mygray}
        1&1  & 73.1 & 80.1 \\
        2&5  & 72.2 & 79.5 \\
        3&8  & 71.7 & 79.0 \\
        4&10 & 71.2 & 78.8 \\
        \hline
    \end{tabular}
    }
    }
    \subfloat[{Data Augmentation} \label{table:data_aug}]{
    \resizebox{0.293\textwidth}{!}{
    \setlength\tabcolsep{1.8pt}
    \begin{tabular}{c|c|c||c}
        \hline\thickhline
	&Augmentation & \multicolumn{1}{c||}{DAVIS$_{17}$} & \multicolumn{1}{c}{DAVIS$_{16}$} \\ \cline{3-4}
        \multirow{-2}{*}{\#}&\multirow{-1}{*}{Strategy} & $\mathcal{J}$\&$\mathcal{F}_m$ $\uparrow$  & $\mathcal{J}$\&$\mathcal{F}_m$ $\uparrow$
        \\ \hline\hline
        \rowcolor{mygray}
        1&Random Crop         & 73.1 & 80.1 \\
        2&+Transformation     & 72.9 & 80.0 \\
        3&+Color Jittering~~  & 71.0 & 78.3 \\
        4&+Gaussian Blur~~    & 71.4 & 78.9 \\
        \hline
    \end{tabular}
    }
    }
    \subfloat[{Dynamic Loss} \label{table:dy_loss}]{
    \resizebox{0.229\textwidth}{!}{
    \setlength\tabcolsep{1.8pt}
    \begin{tabular}{c|c|c||c}
        \hline\thickhline
	&Weight & \multicolumn{1}{c||}{DAVIS$_{17}$} & \multicolumn{1}{c}{DAVIS$_{16}$} \\ \cline{3-4}
        \multirow{-2}{*}{\#}&\multirow{-1}{*}{Factor} & $\mathcal{J}$\&$\mathcal{F}_m$ $\uparrow$  & $\mathcal{J}$\&$\mathcal{F}_m$ $\uparrow$
        \\ \hline\hline
        1&0.1  & 71.9 & 79.1 \\
        2&0.5  & 72.6 & 79.6 \\
        \rowcolor{mygray}
        3&1.0  & 73.1 & 80.1 \\
        4&2.0  & 73.0 & 80.0 \\
        \hline
    \end{tabular}
    }
    }
    \subfloat[{Positive Discriminant} \label{table:ablation4}]{
    \resizebox{0.236\textwidth}{!}{
    \setlength\tabcolsep{1.8pt}
    \begin{tabular}{c|c|c||c}
        \hline\thickhline
	&Positive & \multicolumn{1}{c||}{DAVIS$_{17}$} & \multicolumn{1}{c}{DAVIS$_{16}$} \\ \cline{3-4}
        \multirow{-2}{*}{\#}&\multirow{-1}{*}{Radius} & $\mathcal{J}$\&$\mathcal{F}_m$ $\uparrow$  & $\mathcal{J}$\&$\mathcal{F}_m$ $\uparrow$
        \\ \hline\hline
        1&0.01 & 71.5 & 78.4 \\
        \rowcolor{mygray}
	2&0.10 & 73.1 & 80.1 \\
        3&0.30 & 72.8 & 79.9 \\
	4&0.70 & 72.2 & 79.3 \\
	\hline
    \end{tabular}
    }
    }
    \\
    \subfloat[{Searching Dataset \& Positive Radius}\label{table:dataset_r}]{%
    \resizebox{0.45\textwidth}{!}{
    \setlength\tabcolsep{1.8pt}
    \begin{tabular}{c|c|c|c|c|c||c}
        \hline\thickhline
	&Pre-training &Is &Dataset &Best & \multicolumn{1}{c||}{DAVIS$_{17}$} & \multicolumn{1}{c}{DAVIS$_{16}$} \\ \cline{6-7}
        \multirow{-2}{*}{\#} &\multirow{-1}{*}{Dataset} &\multirow{-1}{*}{Video} &\multirow{-1}{*}{Size} &\multirow{-1}{*}{$r$} & $\mathcal{J}$\&$\mathcal{F}_m$ $\uparrow$  & $\mathcal{J}$\&$\mathcal{F}_m$ $\uparrow$
        \\ \hline\hline
	1& YouTube-VOS~\cite{xu2018youtube} & \cmark & 447K & 0.10  & 73.1  & 80.1 \\
        2& MSRA10k~\cite{cheng2014global}  & \xmark & 10K & 0.10 & 71.1  & 78.4 \\
        3& PASCAL VOC~\cite{everingham2010pascal}  & \xmark &17K & 0.10 & 72.0  & 78.7 \\
        \rowcolor{mygray}
        4& COCO~\cite{lin2014microsoft} & \xmark &118K & 0.10 & 73.1 & 80.1 \\
        \hline
    \end{tabular}
    }
    }
    \subfloat[{Space Projection}\label{table:ablation3}]{%
    \resizebox{0.255\textwidth}{!}{
    \setlength\tabcolsep{1.8pt}
    \begin{tabular}{c|c|c||c}
        \hline\thickhline
	&Hidden & \multicolumn{1}{c||}{DAVIS$_{17}$} & \multicolumn{1}{c}{DAVIS$_{16}$} \\ \cline{3-4}
        \multirow{-2}{*}{\#}&\multirow{-1}{*}{Dimension} & $\mathcal{J}$\&$\mathcal{F}_m$ $\uparrow$  & $\mathcal{J}$\&$\mathcal{F}_m$ $\uparrow$
        \\ \hline\hline
	1&128   & 72.5  & 79.7 \\
	\rowcolor{mygray}
        2&256   & 73.1  & 80.1 \\
        3&512   & 72.9  & 80.0 \\
        4&1,024 & 72.1  & 78.9 \\
        \hline
    \end{tabular}
    }
    }
    \subfloat[{Baseline Selection} \label{table:ablation5}]{
    \resizebox{0.283\textwidth}{!}{
    \setlength\tabcolsep{1.8pt}
    \begin{tabular}{c|c|c||c}
        \hline\thickhline
		&Semantic & \multicolumn{1}{c||}{DAVIS$_{17}$} & \multicolumn{1}{c}{DAVIS$_{16}$} \\ \cline{3-4}
            \multirow{-2}{*}{\#}&\multirow{-1}{*}{Model} & $\mathcal{J}$\&$\mathcal{F}_m$ $\uparrow$  & $\mathcal{J}$\&$\mathcal{F}_m$ $\uparrow$
            \\ \hline\hline
            \rowcolor{mygray}
            1& MoCo~\cite{he2020momentum}    & 67.1$\rightarrow$\color{mygreen}{\textbf{73.1}} & 72.5$\rightarrow$\color{mygreen}{\textbf{80.1}} \\ 
		2& PixPro~\cite{xie2021propagate}  & 59.2$\rightarrow$\color{mygreen}{\textbf{70.0}} & 62.8$\rightarrow$\color{mygreen}{\textbf{77.7}} \\
            3& SimSiam~\cite{chen2020simsiam} & 67.0$\rightarrow$\color{mygreen}{\textbf{71.5}} & 72.7$\rightarrow$\color{mygreen}{\textbf{79.3}} \\
            4& DINO~\cite{caron2021emerging}    & 65.8$\rightarrow$\color{mygreen}{\textbf{70.9}} & 72.4$\rightarrow$\color{mygreen}{\textbf{79.6}} \\
		\hline
    \end{tabular}
    }
    }
\end{table*}

\noindent\textbf{Analysis of Critical Modules} (Eq.~\eqref{eq:6}). To analyze the respective contributions of proposed core components (\ie, static and dynamic visual correspondence), we perform a thorough ablation study, shown in Table~\ref{table:ablation1}. Compared to the baseline (\#1, MoCo~\cite{he2020momentum}), adding the two visual correspondence modules proposed above improves the performance by \textbf{3.5\%} (\# 2) and \textbf{2.8\%} (\#3) $\mathcal{J}\&\mathcal{F}_{m}$ on DAVIS$_{17}$, respectively. For the single object dataset DAVIS$_{16}$, two modules outperform the baseline by \textbf{4.9\%} (\#2) and \textbf{4.5\%} (\#3) in $\mathcal{J}\&\mathcal{F}_{m}$ score. The results in the ablation study show that learning static and dynamic visual correspondence (\#4, HVC) facilitates semantic self-supervised feature representations.

\noindent\textbf{Step between Images} (\S\ref{sec:3.2}). To investigate the impact of the amount of training data on HVC, we set four step sizes: 1, 5, 8, and 10. Step sizes 5, 8 and 10 mean that the COCO~\cite{lin2014microsoft} dataset (\textbf{118K}) is sampled at corresponding image intervals, \ie, 5, 8 and 10 times less data volume. As shown in Table \ref{table:ablation2}, our method at step size 5 decreases only 0.9\% and 0.6\% $\mathcal{J}\&\mathcal{F}_{m}$ on DAVIS$_{16}$ and DAVIS$_{17}$, respectively, compared to step size 1. Remarkably, even with only \textbf{24K} images for training, HVC still outperforms the top-performing LIIR~\cite{li2022liir} that uses the YouTube-VOS$_{18}$ dataset with all frames (\textbf{470K}).

\noindent\textbf{Data Augmentation} (\S\ref{sec:3.3}). The ablation study, as presented in Table~\ref{table:data_aug}, thoroughly investigates the effects of diverse data augmentation strategies on the efficacy of HVC. Singular random cropping is the most effective strategy, echoing findings from preceding relevant studies~\cite{xu2021rethinking,hu2022sfc}. Notably, adding color jittering or Gaussian blur significantly deteriorates model performance. Furthermore, additional transformations (\eg, affine and perspective) have little influence on performance yet incur increased computational overhead.

\noindent\textbf{Dynamic Loss} (Eq.~\eqref{eq:6}). As shown in Table~\ref{table:dy_loss}, we conduct an ablation study, examining the influence of the dynamic loss weight factor on performance. We observe that HVC achieves optimal results when $\alpha$ equals 1.0. Conversely, if $\alpha$ is set below 1.0, there is a notable decrease in performance. The empirical evidence suggests that assigning equivalent significance to both static and dynamic losses fosters a harmonized feature representation within our model.

\begin{figure}[t]
    \begin{center}
	\includegraphics[width=1\linewidth]{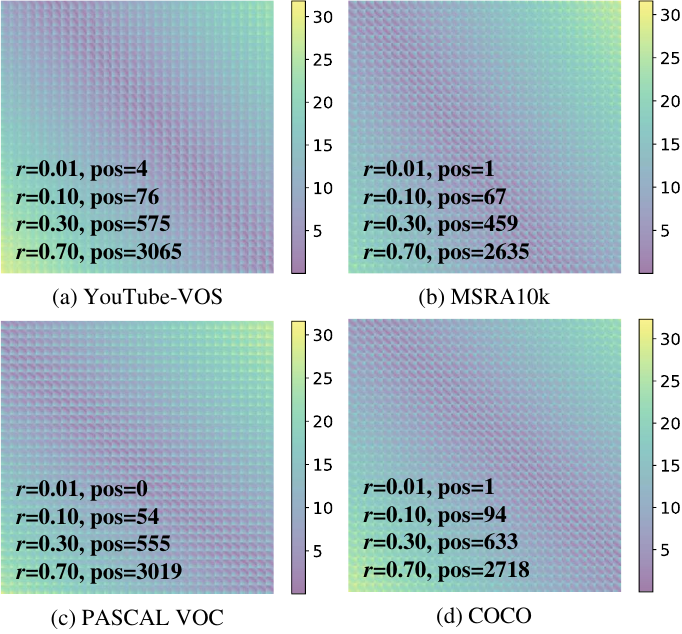}
    \end{center}
    \vspace{-10pt}
    \caption{\textbf{Visualization for feature distances} (\S\ref{sec:abs}) on four crop views from different datasets. `pos' represents the number of positive samples.}\label{fig:pos_dist}
    \vspace{-10pt}
\end{figure}

\noindent\textbf{Positive Discriminant} (Eq.~\eqref{eq:4}). In Table~\ref{table:ablation4}, we investigate the impact of different positive radius on performance. The positive discriminants are determined by the coordinates of the two crop views, providing supervision for visual correspondence learning.
Fig.~\ref{fig:pos_dist} illustrates the feature distances between crop views from different datasets and includes the number of positive samples corresponding to each $r$-value.
A larger positive radius leads to an increase in too many wrong positive pairs, resulting in performance degradation. Conversely, a smaller radius makes the model overly strict and thus reduces the number of positive examples, which leads to more performance reduction. When the positive radius is set to 0.1, HVC reaches the best results on both DAVIS$_{17}$ and DAVIS$_{16}$ (\ie, \textbf{73.1\%} and \textbf{80.1\%} over $\mathcal{J}\&\mathcal{F}_{m}$).

\noindent\textbf{Searching Dataset \& Positive Radius} (Eq.~\eqref{eq:4}). To investigate the effect of different datasets on HVC, we perform pre-training on two additional image-level datasets (\ie, MSRA10k~\cite{cheng2014global}, PASCAL VOC~\cite{everingham2010pascal}) and a widely-used video-level dataset, \ie, YouTube-VOS~\cite{xu2018youtube} (treating frames as separate images).
As shown in Table~\ref{table:dataset_r}, the performance of our method trained on COCO is comparable to the video-level dataset YouTube-VOS$_{18}$, despite COCO (118K) having only a quarter of the number of images as YouTube-VOS (470K). Meanwhile, our method achieves competitive performance on PASCAL VOC (17K) and MSRA10k (10K), which have \textbf{28} and \textbf{47} times fewer images than YouTube-VOS$_{18}$. Noteworthy, HVC only requires \textbf{0.5h} to complete pre-training on PASCAL VOC~\cite{everingham2010pascal} and has a score of \textbf{72.0\%} $\mathcal{J}\&\mathcal{F}_{m}$, which is quite close to the state-of-the-art LIIR~\cite{li2022liir} (72.1\% $\mathcal{J}\&\mathcal{F}_{m}$) trained on YouTube-VOS~\cite{xu2018youtube} for 12h.
Combining the results of the pre-trained models on each data, we find that setting $r$ to 0.1 always leads to the best results.

\noindent\textbf{Space Projection} (Eq.~\eqref{eq:flow}). To elucidate the efficacy of our space projection module, we report the ablation results for different hidden dimensions, as shown in Table~\ref{table:ablation3}. Setting the hidden dimension to 256 yields a highly competitive performance with low computational cost. Employing a larger dimension incurs extra computational overhead during the training process, leading to diminished efficiency with minimal benefit to model performance. The reason is that the space projection module, as the projector head and predictor head for dense features from the backbone, is limited by the dimension of the original feature channels. Hence, the additional complexity introduced by the larger dimensions might not be effectively applied by the model, especially given the amount and diversity of training data.

\noindent\textbf{Baseline Selection} (\S\ref{sec:3.3}) Self-supervised methods designed for image recognition have rich semantic representations. To study the effect of incorporating varied methods (\ie, MoCo~\cite{he2020momentum}, PixPro~\cite{xie2021propagate}, SimSiam~\cite{chen2020simsiam}, and DINO~\cite{caron2021emerging}) of semantic self-supervision with the proposed approach, we perform the ablation experiments shown in Table~\ref{table:ablation5}. The initial scores of MoCo and SimSiam are very close. Eventually, MoCo achieves better performance, likely due to the architectural similarities between our HVC and MoCo~\cite{he2020momentum}.

\begin{figure}[t]
    \begin{center}
        \includegraphics[width=0.95\linewidth]{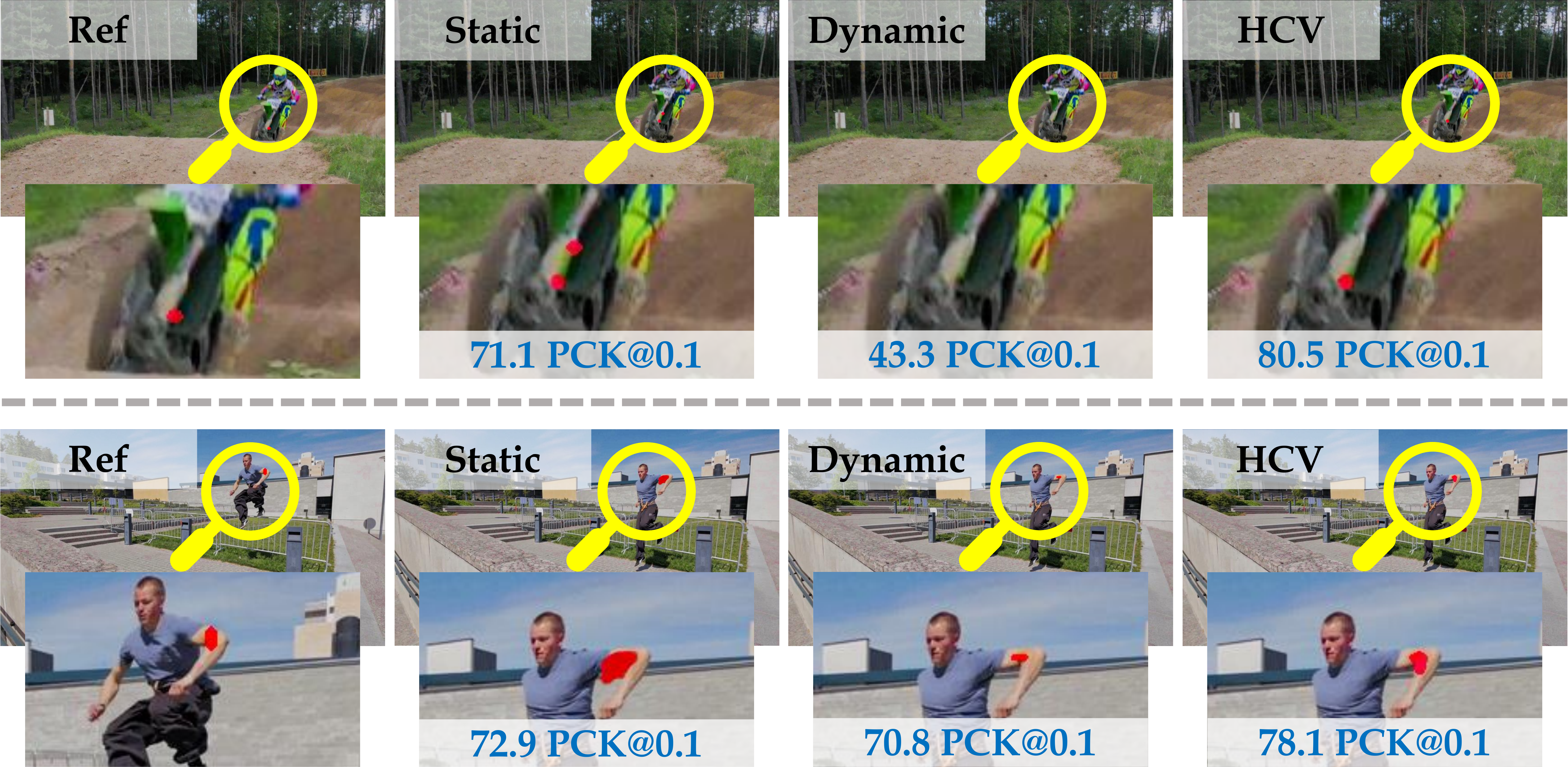}
    \end{center}
    \vspace{-6pt}
    \caption{\textbf{Qualitative results of key modules} (\S\ref{sec:abs}) for semantic consistency. Given a reference scribbling of the first frame, infer the scribble area in later frames and mark PCK@0.1. Zoomed-in view for best.}\label{fig:pt}
\end{figure}

\begin{figure}[t]
    \begin{center}
	\includegraphics[width=1\linewidth]{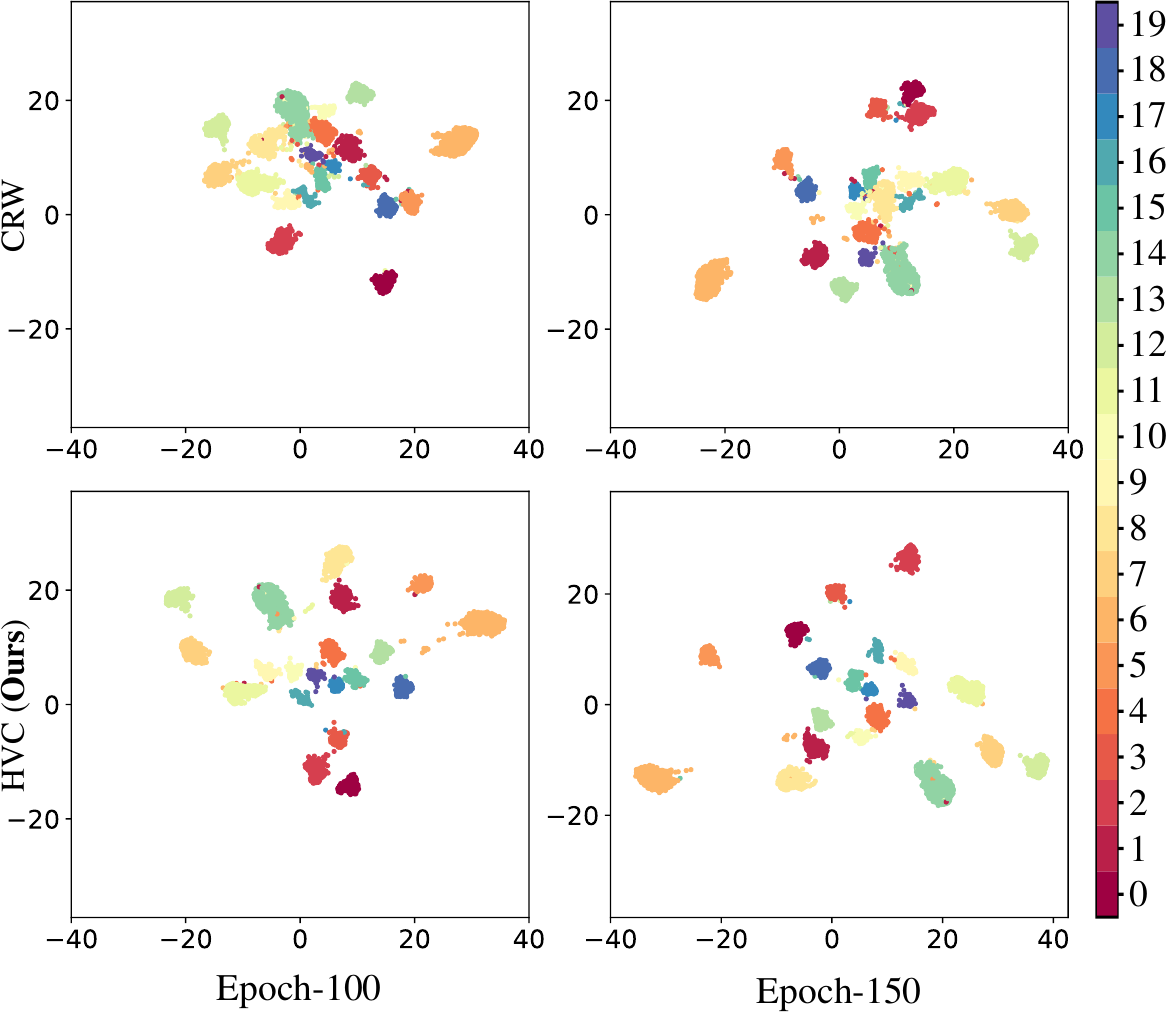}
    \end{center}
    \vspace{-10pt}
    \caption{\textbf{Visualization for feature embedding} (\S\ref{sec:feat_embed}) on PASCAL VOC~\cite{everingham2010pascal} \texttt{test-set}. Top: the feature embedding of CRW~\cite{jabri2020space}; Bottom: the feature embedding of our HVC. Zoomed-in view for best.}\label{fig:fe}
    \vspace{-6pt}
\end{figure}

\begin{table}[t]
    \caption{\textbf{Comparison results} (\S\ref{sec:eff_anal}) of computational efficiency. \# and $\Delta t$ denote training iterations and training time, respectively.}\label{table:efficiency}
    \centering
        \resizebox{\columnwidth}{!}{
        \setlength\tabcolsep{6.5pt}
        \renewcommand\arraystretch{1.0}
        \begin{tabular}{ccccc}
            \hline\thickhline
            Method &Dataset & \# & $\Delta t$ & GPU Memory \\ \hline\hline
            ContrastCorr~\cite{wang2021contrastive}   & C + TN   & 0.4M  & 1d   & 44GB     \\
            CRW~\cite{jabri2020space}                 & Kinetics & 2M    & 7d   & 22GB     \\
            DUL~\cite{dense2021}                      & Kinetics & 0.3M  & 2d   & 12GB     \\
            MAST~\cite{lai2020mast}                   & YT       & 2M    & 20h  & 22GB     \\
            LIIR~\cite{li2022liir}                    & YT       & 2M    & 12h  & 48GB     \\
            DUL~\cite{dense2021}                      & YT       & 0.1M  & 16h  & 12GB     \\
            SFC~\cite{hu2022sfc}                      & YT       & 0.1M  & 25h  & 20GB     \\
            \hdashline
            \textbf{HVC-YT~(ours)} & YT       & 0.1M  & 7h   & 16GB     \\ 
            \rowcolor{rowgray} \textbf{HVC~(ours)}    & C        & 0.05M & 2h   & 16GB     \\ 
            \hline
            \end{tabular}
            }
\end{table}

\begin{table}[t]
    \caption{\textbf{Quantitative results for video object segmentation} (\S\ref{sec:add_data}) on DAVIS$_{17}$~\cite{pont2017} \texttt{test-set}.}\label{table:davis-17test}
    \centering
        \resizebox{\columnwidth}{!}{
	\setlength\tabcolsep{4.6pt}
	\renewcommand\arraystretch{1.1}
	\begin{tabular}{ccccccc}
            \hline\thickhline
            Method & Sup. & $\mathcal{J}\&\mathcal{F}_{m}$ $\uparrow$ & $\mathcal{J}_{m}$ $\uparrow$ & $\mathcal{J}_{r}$ $\uparrow$ & $\mathcal{F}_{m}$ $\uparrow$ & $\mathcal{F}_{r}$ $\uparrow$ \\ \hline\hline
		CRW~\cite{jabri2020space}     & \xmark  & 55.9 & 52.3 & -    & 59.6 & - \\
		VFS~\cite{xu2021rethinking}   & \xmark  & 57.3 & 53.1 & -    & 61.6 & - \\
		SCC~\cite{son2022contrastive} & \xmark  & 59.9 & 55.9 & 63.9 & 64.0 & 72.7 \\
            \hdashline
		\textbf{HVC-YT~(ours)}           & \xmark  & \textbf{61.8} & 57.3 & 64.7 & \textbf{66.3} & \textbf{74.9} \\ 
  		\rowcolor{rowgray} \textbf{HVC (ours)}           & \xmark  & 61.7 & \textbf{57.5} & \textbf{65.4} & 66.0 & 74.8 \\ 
		\hline
	\end{tabular}
	}
\end{table}

\begin{figure*}
    \begin{center}
	\includegraphics[width=1\linewidth]{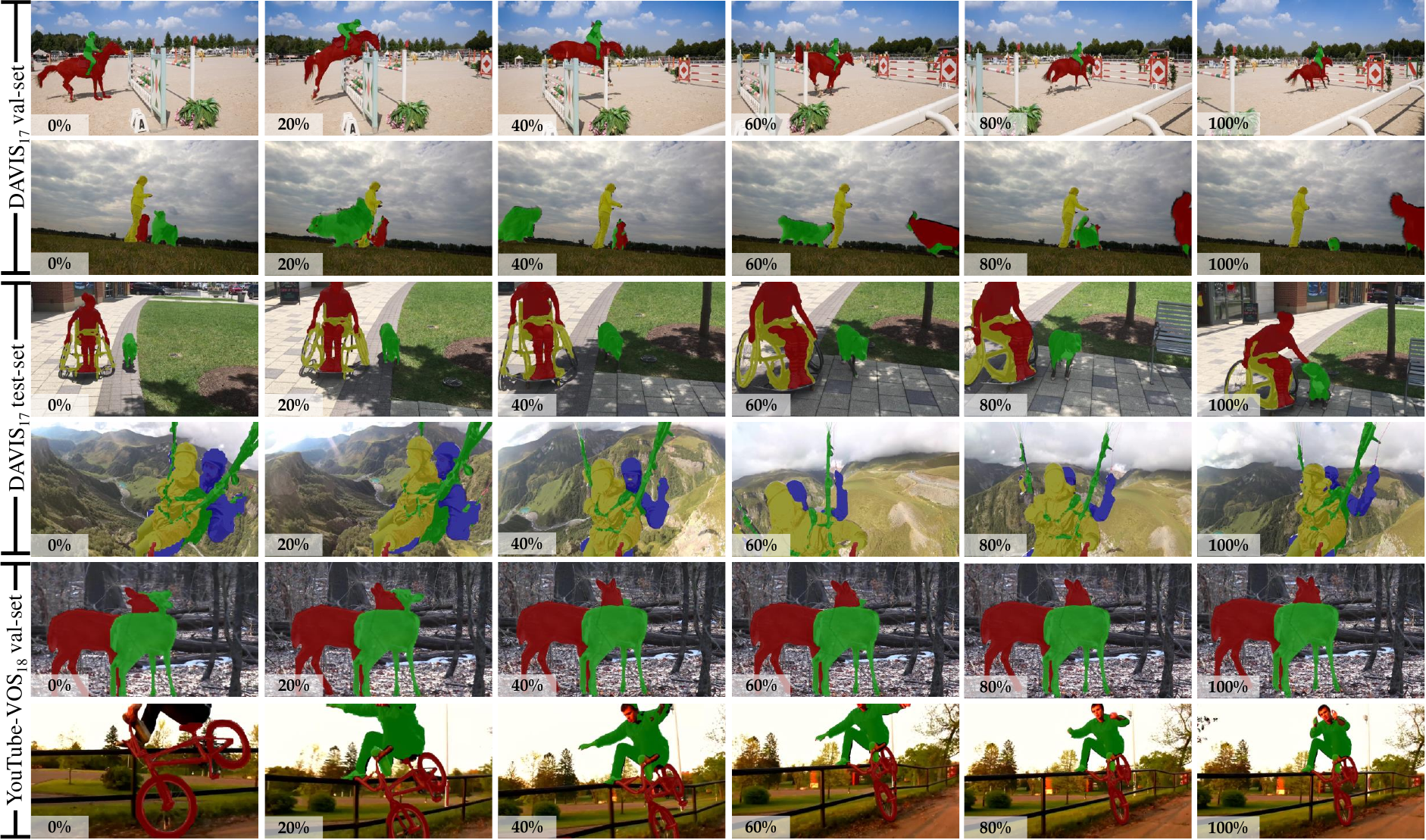}
    \end{center}
    \vspace{-8pt}
    \caption{\textbf{Additional qualitative results for video object segmentation} (\S\ref{sec:add_data}). The video progress status is shown from 0\% to 100\% with an interval of 20\%. The test video sequences are from the DAVIS$_{17}$ \texttt{val-set}, DAVIS$_{17}$~\cite{pont2017} \texttt{test-set}, and YouTube-VOS$_{18}$~\cite{xu2018youtube} \texttt{val-set}. The feature representation capability of our proposed HVC yields delicate segmentation results in multiple challenging datasets.}\label{fig:add_vis}
    \vspace{-6pt}
\end{figure*}

\subsection{Further Analysis} \label{sec:FurExp}
\subsubsection{Efficiency} \label{sec:eff_anal}
Achieving optimal video feature representation typically demands extensive video data, leading to prolonged training time and substantial resource consumption. In contrast, our approach utilizes only images to train feature representations while maintaining static-dynamic consistency, striking a balance between performance and complexity. Table~\ref{table:efficiency} presents a comprehensive analysis of the training efficiency of HVC and other methods, considering factors such as training iterations, time, and GPU memory usage. Remarkably, our method achieves state-of-the-art performance using only static image data as training data while requiring little training time (\textbf{2h}) and GPU memory resources (\textbf{16GB}).

\subsubsection{Visualization of Feature Tracking and Embedding} \label{sec:feat_embed}
In this experiment, we delve deeper into the effectiveness of the learned feature representation by the proposed model on label propagation and image recognition tasks. \textbf{1)} As shown in Fig.~\ref{fig:pt}, we replace the first frame annotation with scribbling and subsequently infer the propagation results of this reference mask in the following frames. The learned correspondence by HVC is more detail-oriented, resulting in superior performance compared to its static and dynamic variants. \textbf{2)} Fig.~\ref{fig:fe} depicts low-dimensional embedding visualization of HVC on PASCAL VOC~\cite{everingham2010pascal}. Our approach has better feature discrimination than CRW~\cite{jabri2020space}. Employing the UMAP~\cite{mcinnes2018umap} training mode, the projection of the categories becomes more accurate as the training proceeds. These findings demonstrate that the learned feature representation is highly effective, even in recognition tasks (non-dense tasks).

\subsubsection{Experiments on Additional VOS Datasets} \label{sec:add_data}
\noindent\textbf{Quantitative Results.} We further expand the evaluation of HVC and compare it with state-of-the-art self-supervised methods on more challenging datasets. We present the qualitative results of our method on DAVIS$_{17}$\cite{pont2017} \texttt{test-set} in Table~\ref{table:davis-17test}. This dataset consists of 30 video sequences with a higher occlusion frequency compared to DAVIS$_{17}$ \texttt{val-set}. HVC achieves an impressive score of \textbf{61.7\%} $\mathcal{J}\&\mathcal{F}_{m}$, surpassing existing self-supervised methods. Specifically, our approach shows a \textbf{1.8\%} improvement over SCC~\cite{son2022contrastive}. Furthermore, we evaluate HVC on the YouTube-VOS$_{19}$ version~\cite{yang2019video}, which includes 507 validation videos. The results presented in Table~\ref{table:ytb-2019}, showing that HVC significantly outperforms existing methods in terms of performance.

\noindent\textbf{Qualitative Results.} 
To further showcase the efficacy of our self-supervised approach in video object segmentation, we provide supplementary qualitative results. As illustrated in Fig.~\ref{fig:add_vis}, we present the qualitative outcomes generated from mask propagation using HVC on three datasets: DAVIS$_{17}$\cite{pont2017} \texttt{val-set}, DAVIS$_{17}$\cite{pont2017} \texttt{test-set}, and YouTube-VOS$_{18}$~\cite{xu2018youtube} \texttt{val-set}. The application of learned feature representations, facilitated by our HVC, yields notable performance across a spectrum of complex video sequences.

\begin{table}[t]
    \caption{\textbf{Quantitative results for video object segmentation} (\S\ref{sec:add_data}) on YouTube-VOS$_{19}$~\cite{yang2019video}
    	\vspace{-0.3cm}
    	 \texttt{val-set}.}\label{table:ytb-2019}
    \centering
        \resizebox{\columnwidth}{!}{
        \setlength\tabcolsep{5pt}
        \renewcommand\arraystretch{1.125}
        \begin{tabular}{ccccccc}
            \hline\thickhline
            & & & \multicolumn{2}{c}{Seen} & \multicolumn{2}{c}{Unseen} \\ \cline{4-7}
            \multirow{-2}{*}{Method} & \multirow{-2}{*}{Sup.} & \multirow{-2}{*}{Mean} & $\mathcal{J}_{m}$ $\uparrow$ & $\mathcal{F}_{m}$ $\uparrow$ & $\mathcal{J}_{m}$ $\uparrow$ & $\mathcal{F}_{m}$ $\uparrow$\\ \hline\hline
            Colorization~\cite{vondrick2018tracking} & \xmark & 39.0 & 43.3 & 38.2 & 36.6 & 37.5 \\
            CorrFlow~\cite{lai2019self} & \xmark & 47.0 & 51.2 & 46.6 & 44.5 & 45.9 \\
            MAST~\cite{lai2020mast} & \xmark & 64.9 & 64.3 & 65.3 & 61.5 & 68.4 \\
            \hdashline
            \textbf{HVC-YT~(ours)} & \xmark & 71.3 & 69.2 & 70.9 & 67.8 & 77.2 \\ 
            \rowcolor{rowgray} \textbf{HVC~(ours)} & \xmark & \textbf{71.6} & \textbf{69.5} & \textbf{71.2} & \textbf{68.2} & \textbf{77.4} \\ 
            \hline
	\end{tabular}
	}
\end{table}

\begin{table}[t]
    \caption{\textbf{Quantitative results of proposed pseudo-dynamic module} (\S\ref{sec:dynamic_flow}) on DAVIS$_{17}$~\cite{pont2017} \texttt{val-set} and YouTube-VOS$_{18}$~\cite{xu2018youtube} \texttt{val-set}. `$\Delta$' means the estimated gap between our HVC and off-the-shelf models.}\label{table:pseudo-dynamic}
    \vspace{-0.3cm}
    \centering
    \resizebox{\columnwidth}{!}{
    \setlength\tabcolsep{5pt}
    \renewcommand\arraystretch{1.1}
    \begin{tabular}{c|c|c||c}
        \hline\thickhline
	& & \multicolumn{1}{c||}{DAVIS$_{17}$} & \multicolumn{1}{c}{YouTube-VOS$_{18}$} \\ \cline{3-4}
        \multirow{-2}{*}{Off-the-Shelf} & \multirow{-2}{*}{Direction} & EPE $\downarrow$ & EPE $\downarrow$ \\ \hline\hline
                                                          & Forward  & 6.52 & 13.18 \\
        \multirow{-2}{*}{$\Delta$HVC-RAFT~\cite{teed2020raft}}  & Backward & 5.71 & 13.13 \\
                                                                  & Forward  & 5.10 & 6.88 \\
        \multirow{-2}{*}{$\Delta$HVC-PWC~\cite{sun2018pwc}}  & Backward & 4.49 & 6.83 \\
	\hline
    \end{tabular}
    }
\end{table}

\subsubsection{Pseudo-Dynamic Signals}\label{sec:dynamic_flow}
Our method employs two streamlined convolutional layers (see \S\ref{sec:3.2.2} for details) to calculate pseudo-dynamic signals, resembling optical flows, amongst cropped image views. This minimalist design circumvents overcomplication of the model while maintaining optimal performance.

\noindent\textbf{Quantitative Analysis.}
We utilize the off-the-shelf models (\eg, RAFT~\cite{teed2020raft} and PWC~\cite{sun2018pwc}) for optical flow estimates on the DAVIS~\cite{pont2017} and YouTube-VOS$_{18}$~\cite{xu2018youtube} validation sets as a ground truth benchmark. We then apply our module $\mathcal{F}_{\rm{pseudo}}$ (Eq.~\eqref{eq:3}) to these datasets to produce dynamic signal estimations. As shown in Table~\ref{table:pseudo-dynamic}, the gap in the End-Point Error (EPE) metric between our self-supervised method and these supervised models is relatively small, signifying that our pseudo-dynamic signal estimations are competitive.

\begin{figure}[t]
    \begin{center}
        \includegraphics[width=1\linewidth]{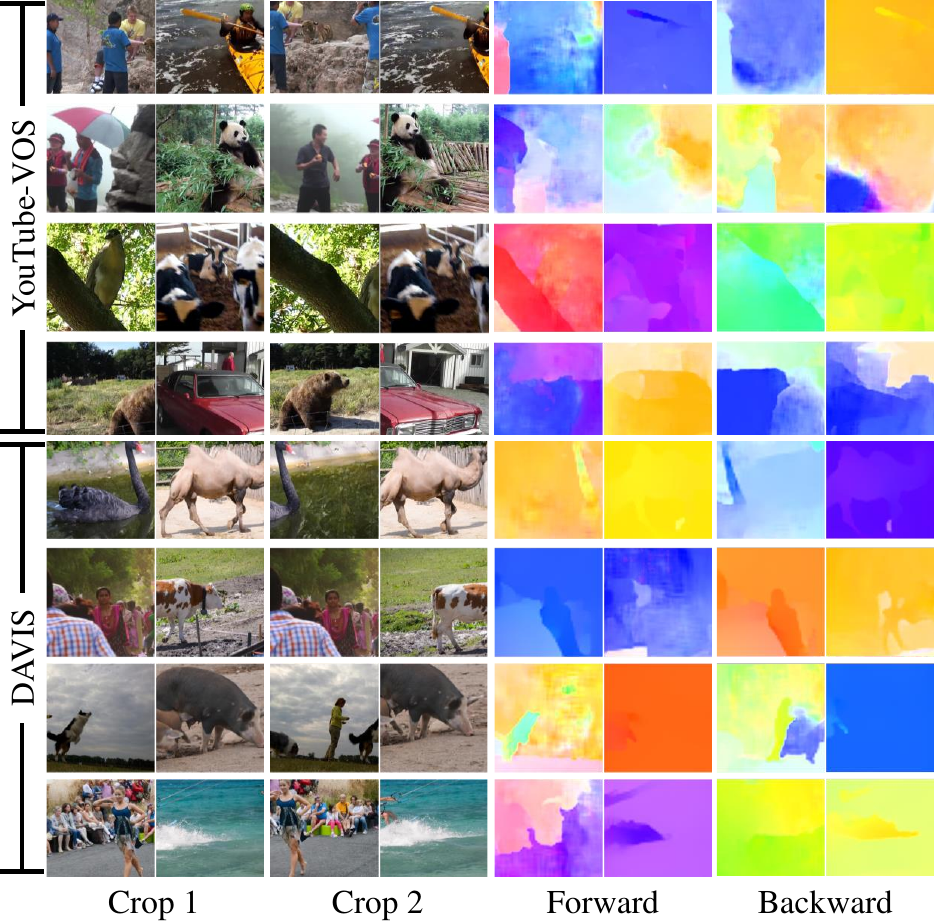}
    \end{center}
    \vspace{-8pt}
    \caption{\textbf{Pseudo-dynamic signal visualization} (\S\ref{sec:dynamic_flow}) on YouTube-VOS~\cite{xu2018youtube} (Top) and DAVIS~\cite{perazzi2016benchmark} (Bottom). `Forward' and `Backward' indicate the forward and backward pseudo-dynamic signals of `Crop1' and `Crop2' from one image, respectively. Zoomed-in view for best.}\label{fig:dynamic_flow}
\end{figure}

\noindent\textbf{Qualitative Analysis.}
Following the standardized visualization method proposed in~\cite{baker2011database}, we transform the color space from $2 \times H \times W$ to $3 \times H \times W$. Fig.~\ref{fig:dynamic_flow} illustrates the visualization of pseudo-dynamic signals in use on YouTube-VOS~\cite{xu2018youtube} and DAVIS~\cite{pont2017} datasets. Mirroring the functionality of inter-frame optical flow, our proposed pseudo-dynamic signals between cropped views effectively capture motion cues to maintain dynamic consistency. This assures static-dynamic coherence which in turn enhances the performance of self-supervised feature representation models initially trained on image data.

\section{Limitation Analysis and Discussion} \label{sec:limit}
Our study presents a hybrid static-dynamic visual correspondence approach that mitigates the influence of uncertain self-supervised signals impeding model convergence.
However, it encounters challenges inherent to VOS, such as modeling long-term video sequences and dealing with significant appearance changes caused by occlusion or deformation.
Especially, the lack of bespoke pre-training data for intricate scenarios amplifies these challenges.

\noindent\textbf{Long-Term Correspondence.} Drawing insights from \cite{haller2021iterative,haller2019spacetime} of leveraging motion chains, we realize an opportunity to refine our approach. Integrating concepts from their iterative knowledge exchange framework can potentially enrich object discovery over vast temporal extents. This can offer a robust solution to a noted limitation of our method: visual correspondences merely between adjacent frames.

\noindent\textbf{Pseudo-Dynamic Generation.} Recently, the generation of pseudo-frames from static images has become feasible, aiding video representation learning via readily available diffusion techniques~\cite{rombach2022sd,zhang2023controlnet,tang2023mvdiffusion,ren2024move}. Thus, our future research will focus on scalable diffusion techniques in self-supervised VOS, aimed at generating controllable pseudo-frames and enhancing the robustness of VOS in restricted settings.

\section{Conclusion}
In this paper, we presented HVC, an elegant yet highly efficient self-supervised approach for video object segmentation. In contrast to previous approaches that rely on large amounts of training data or complex frame reconstruction methods, our approach achieved superior performance and efficiency by leveraging an image-based self-supervised representation learning paradigm, such as MoCo~\cite{he2020momentum}. By simulating the dynamic signal presented in video data through image-cropped views, HVC ensured static-dynamic consistency in hybrid visual correspondence learning for VOS. Across multiple benchmark datasets, HVC, trained using only static images, surpassed existing state-of-the-art self-supervised methods that were trained on videos. Moreover, our approach demonstrated the applicability of image-based correspondence learning~\cite{xie2021propagate,caron2021emerging} for video segmentation tasks. We hope the proposed method provides an alternative perspective for self-supervised video object segmentation, opening up new possibilities for utilizing static images in the learning process.

{\small
\bibliographystyle{IEEEtran}

\bibliography{egbib}
}

\end{document}